\documentclass[twoside,11pt]{article}
\usepackage{jmlr2e}
\usepackage{amsmath}
\usepackage[USenglish]{babel} %francais, polish, spanish, ...
\usepackage{bm}% bold math
\usepackage{setspace}
\usepackage[]{natbib}
\usepackage{graphicx}
\usepackage{algorithm}
\usepackage{algorithmic}
\usepackage{multirow}
\usepackage{url}
\usepackage{array}
\usepackage{colortbl}
\usepackage{epstopdf}
\ShortHeadings{Fast SVM training using approximate extreme points}{Nandan, Khargonekar, and Talathi}
\firstpageno{1}

\begin{document}

\title{Fast SVM Training Using Approximate Extreme Points}

\author{\name Manu Nandan \email mnandan@ufl.edu \\
       \addr Department of Computer and Information Science and Engineering\\
       University of Florida\\
       Gainesville, FL 32611, USA
       \AND
       \name Pramod P. Khargonekar \email ppk@ece.ufl.edu \\
       \addr Department of Electrical and Computer Engineering\\
       University of Florida,\\
       Gainesville, FL 32611, USA
       \AND
       \name Sachin S. Talathi \email talathi@ufl.edu \\
       \addr Department of Pediatrics, Division of Neurology\\  
						 Department of Biomedical Engineering\\
						 Department of Neuroscience\\ 
       University of Florida\\
       Gainesville, FL 32611, USA}

\editor{}

\maketitle

\begin{abstract}%
Applications of non-linear kernel Support Vector Machines (SVMs) to  large datasets is seriously hampered by its excessive training time. We propose a modification, called the approximate extreme points support vector machine (AESVM), that is aimed at overcoming this burden. Our approach relies on conducting the SVM optimization over a carefully selected subset, called the representative set, of the training dataset. We present analytical results that indicate the similarity of AESVM and SVM solutions. A linear time algorithm based on convex hulls and extreme points is used to compute the representative set in kernel space. Extensive computational experiments on nine datasets compared AESVM to LIBSVM \citep{LIBSVM}, CVM \citep{Tsang05} , BVM \citep{Tsang07}, LASVM \citep{Bordes05}, $\text{SVM}^{\text{perf}}$ \citep{Joachims09}, and the random features method \citep{rahimi07}. Our AESVM implementation was found to train much faster than the other methods, while its classification accuracy was similar to that of LIBSVM in all cases. In particular, for a seizure detection dataset, AESVM training was almost $10^3$ times faster than LIBSVM and LASVM and more than forty times faster than CVM and BVM.  Additionally, AESVM also gave competitively fast classification times.  
\end{abstract} 


\section{Introduction}

Several real world applications require solutions of classification problems on large datasets. Even though SVMs are known to give excellent classification results, their application to problems with large datasets is impeded by the burdensome training time requirements. Recently, much progress has been made in the design of fast training algorithms \citep{Fan08,Shwartz11} for SVMs with the linear kernel (linear SVMs). However, many applications require SVMs with non-linear kernels for accurate classification. Training time complexity for SVMs with non-linear kernels is typically quadratic in the size of the training dataset \citep{Shalev08}. The difficulty of the long training time is exacerbated when grid search with cross-validation is used to derive the optimal hyper-parameters, since this requires multiple SVM training runs. Another problem that sometimes restricts the applicability of SVMs is the long classification time. The time complexity of SVM classification is linear in the number of support vectors and in some applications the number of support vectors is found to be very large \citep{Guo05}.

In this paper, we propose a new approach for fast SVM training. Consider a two class dataset of $N$ data vectors, ${\bf X} = \{{\bf x}_i: \; {\bf x}_i \in \mathbb{R}^D, i = 1,2,...,N\}$, and the corresponding target labels ${\bf Y} = \{y_i:\; y_{i}\in[-1, 1], i = 1,2,...,N\}$. The SVM primal problem can be represented as the following unconstrained optimization problem \citep{Teo10,Shwartz11}:
\begin{align} 
&\underset{\mathbf{w},b}{\mathit{\min}}\;{F}_1(\mathbf{w} ,b) = \frac{1}{2}\|{\mathbf{w}}\|^2 + \frac{C}{N}\overset{N}{\underset{i = 1}\sum}l(\mathbf{w},b,\phi({\bf x}_i)) \label{eq:SVC_primal2} \\
&\text{where } l(\mathbf{w},b,\phi({\bf x}_i)) = max\{0, 1- y_{i}(\mathbf{w} ^T \phi({\bf x}_i) + b)\}, \forall {\bf x}_i \in {\bf X} \nonumber\\
&\hspace{30pt}\text{and }\phi: \mathbb{R}^D \rightarrow \mathbb{H}, b \in \mathbb{R} \text{, and } {\bf w} \in \mathbb{H} \text{, a Hilbert space} \nonumber
\end{align}
Here $l(\mathbf{w},b,\phi({\bf x}_i))$ is the hinge loss of ${\bf x}_i$. Note that SVM formulations where the penalty parameter $C$ is divided by $N$ have been used extensively \citep{Scholkopf00,Franc08,Joachims09}. These formulations enable better analysis of the scaling of $C$ with $N$ \citep{Joachims06}. The problem in (\ref{eq:SVC_primal2}) requires  optimization over $N$ variables. In general, for SVM training algorithms the training time will reduce if the size of the training dataset is reduced.

{\em In this paper, we present an alternative to (\ref{eq:SVC_primal2}), called approximate extreme points support vector machines (AESVM), that requires optimization over only a subset of the training dataset.} The AESVM formulation is:
\begin{align}
&\underset{\mathbf{w},b}{\mathit{\min}}\;{F}_2(\mathbf{w} ,b) = \frac{1}{2}\|{\mathbf{w}}\|^2 + \frac{C}{N}\overset{M}{\underset{t = 1}\sum}\beta_t l(\mathbf{w},b,\phi({\bf x}_t))  \label{eq:AESVM_primal2}\\
&\text{where } {\bf x}_t \in {\bf X}^*,{\bf w} \in \mathbb{H}\text{, and } b \in \mathbb{R} \nonumber
\end{align}
Here  $M$ is the number of vectors in the selected subset of ${\bf X}$, called the representative set ${\bf X}^*$. The constants $\beta _t$ are defined in (\ref{eq:betaDef}). We will prove in Section \ref{sec:prop} that:

\begin{itemize}
\item ${F}_1 (\mathbf{w}_1 ^*, b_1 ^*) - {F}_2 (\mathbf{w}_2 ^*, b_2 ^*) \le C\sqrt{C\epsilon}$, where $(\mathbf{w}_1 ^*, b_1 ^*)$ and $(\mathbf{w}_2 ^*, b_2 ^*)$ are the solutions of (\ref{eq:SVC_primal2}) and (\ref{eq:AESVM_primal2}) respectively
\item Under the assumptions given in corollary 4, ${F}_1 (\mathbf{w}_2 ^*, b_2 ^*) - {F}_1 (\mathbf{w}_1 ^*, b_1 ^*) \le 2C\sqrt{C\epsilon}$
\item The AESVM problem minimizes an upper bound of a low rank Gram matrix approximation of the SVM objective function
\end{itemize}

Based on these results we claim that solving the problem in (\ref{eq:AESVM_primal2}) yields a solution close to that of (\ref{eq:SVC_primal2}). As a by-product of the reduction in size of the training set, AESVM is also observed to result in fast classification. Considering that the representative set will have to be computed several times if grid search is used to find the optimum hyper-parameter combination, we also propose fast algorithms to compute ${\bf Z}^*$. In particular, we present an algorithm of time complexity $O(N)$ and an alternative algorithm of time complexity $O(N\; \log_2 \frac{N}{P})$ to compute ${\bf Z}^*$, where $P$ is a predefined large integer. 

The main contributions of this work can be summarized as follows:
\begin{itemize}
\item {\it Theoretical:} Theorems 1 and 2, and Corollaries 3 to 5 give rationale for the use of AESVM as a computationally less demanding alternative to the SVM formulation.
\item {\it Algorithmic:} The algorithm DeriveRS, described in Section \ref{sec:derRP}, computes the representative set in linear time.
\item {\it Experimental:} Our extensive experiments on nine datasets of varying characteristics, illustrate the suitability of applying AESVM to classification on large datasets.
\end{itemize}

This paper is organized as follows: in Section 2, we briefly discuss recent research on fast SVM training that is closely related to this work. Next, we provide the definition of the representative set and discuss properties of AESVM. In section 4, we present efficient algorithms to compute the representative set and analyze its computational complexity. Section 5 describes the results of our computational experiments. We compared AESVM to the widely used LIBSVM library, core vector machines (CVM), ball vector machines (BVM), LASVM, $\text{SVM}^{\text{perf}}$, and the random features method by \citet{rahimi07}. Our experiments used eight publicly available datasets and a data set on EEG from an animal model of epilepsy \citep{Talathi08,Nandan10}. We conclude with a discussion of the results of this paper in Section 6.

\section{Related Work} \label{sec:relWork}

Several methods have been proposed to efficiently solve the SVM optimization problem. SVMs require special algorithms, as standard optimization algorithms such as interior point methods \citep{Boyd04,Shwartz11} have large memory and training time requirements that make it infeasible for large datasets. In the following sections we discuss the most widely used strategies to solve the SVM optimization problem. We present a comparison of some of these methods to AESVM in Section \ref{sec:Disc}. SVM solvers can be broadly divided into two categories as described below.
\subsection{Dual optimization} \label{sec:otherDual}
The SVM primal problem is a convex optimization problem with strong duality \citep{Boyd04}. Hence its solution can be arrived at by solving its dual formulation given below:
\begin{align} \label{eq:SVC_dual}
&\underset{\alpha}{\textbf{max }} L_1(\alpha) =  \overset{N}{\underset{i = 1}{\sum}}\alpha_{i} - \frac{1}{2}\overset{N}{\underset{i = 1}{\sum}}\overset{N}{\underset{j = 1}{\sum}}\alpha_{i}\alpha_{j}y_{i}y_{j}K({\bf x}_{i},{\bf x}_{j})\\
&\text{ subject to } 0 \le \alpha_i \le \frac{C}{N} \text{ and } \underset{i = 1}{\overset{N}{\sum}}\alpha_{i}y_i = 0  \nonumber
\end{align}
Here $K({\bf x}_{i},{\bf x}_{j}) = \phi({\bf x}_i)^T\phi({\bf x}_j)$, is the kernel product \citep{Scholkopf01} of the data vectors ${\bf x}_i$ and ${\bf x}_j$, and $\alpha$ is a vector of all variables $\alpha_i$. Solving the dual problem is computationally simpler, especially for non-linear kernels and a majority of the SVM solvers use dual optimization. Some of the major dual optimization algorithms are discussed below.

{\em Decomposition methods} \citep{Osuna97} have been widely used to solve (\ref{eq:SVC_dual}). These methods optimize over a subset of the training dataset, called the `working set', at each algorithm iteration. $\text{SVM}^{light}$ \citep{Joachims99} and SMO \citep{Platt99} are popular examples of decomposition methods. Both these methods have a quadratic time complexity for linear and non-linear SVM kernels \citep{Shalev08}. Heuristics such as shrinking and caching \citep{Joachims99} enable fast convergence of decomposition methods and reduce their memory requirements. LIBSVM \citep{LIBSVM} is a very popular implementation of SMO. A {\em dual coordinate descent} \citep{Hsieh08} SVM solver computes the optimal $\alpha$ value by modifying one variable $\alpha_i$ per algorithm iteration. Dual coordinate descent SVM solvers, such as LIBLINEAR \citep{Fan08}, have been proposed primarily for the linear kernel. 

{\em Approximations of the Gram matrix} \citep{Fine02,Drineas05}, have been proposed to increase training speed and reduce memory requirements of SVM solvers. The Gram matrix is the $N$x$N$ square matrix composed of the kernel products $K({\bf x}_{i},{\bf x}_{j}),\; \forall {\bf x}_i, {\bf x}_j \in {\bf X}$. {\em Training set selection} methods attempt to reduce the SVM training time by optimizing over a selected subset of the training set. Several distinct approaches have been used to select the subset. Some methods use clustering based approaches \citep{Pavlov00} to select the subsets. In \citet{Yu03}, hierarchical clustering is performed to derive a dataset that has more data vectors near the classification boundary than away from it. Minimum enclosing ball clustering is used in \citet{Cervantes08} to remove data vectors that are unlikely to contribute to the SVM training. 

{\em Random sampling} of training data is another approach followed by approximate SVM solvers. \citet{Mangasarian01} proposed reduced support vector machines (RSVM), in which only a random subset of the training dataset is used. They solve a modified formulation of the L2-SVM that  minimizes the $l^2$-norm of $\xi$ instead of its $l^1$-norm. \citet{Bordes05} proposed the LASVM algorithm that uses {\em active selection} techniques to train SVMs on a subset of the training dataset. 

A {\em core set} \citep{Clarkson10} can be loosely defined as the subset of ${\bf X}$ for which the solution of an optimization problem such as (\ref{eq:SVC_dual}) has a solution similar to that for the entire dataset ${\bf X}$. \citet{Tsang05} proved that the L2-SVM is a reformulation of the minimum enclosing ball problem for some kernels. They proposed core vector machine (CVM) that approximately solves the L2-SVM formulation using core sets. A simplified version of CVM called ball vector machine (BVM) was proposed in \citet{Tsang07}, where only an enclosing ball is computed. \citet{Gartner09} proposed an algorithm to solve the L1-SVM problem, by computing the shortest distance between two polytopes \citep{Bennett00} using core sets. However, there are no published results on solving L1-SVM with non-linear kernels using their algorithm. 

Another method used to approximately solve the SVM problem is to map the data vectors into a {\em randomized feature space} that is relatively low dimensional compared to the kernel space $\mathbb{H}$ \citep{rahimi07}. Inner products of the projections of the data vectors are approximations of their kernel product. This effectively reduces the non-linear SVM problem into the simpler linear SVM problem, enabling the use of fast linear SVM solvers. This method is referred as RfeatSVM in the following sections of this document.

\subsection{Primal optimization}
In recent years, linear SVMs are finding increased use in applications with high-dimensional datasets. This has led to a surge in publications on efficient primal SVM solvers, which are mostly used for linear SVMs. To overcome the difficulties caused by the non-differentiability of the primal problem, the following methods are used.

{\em Stochastic sub-gradient descent} \citep{Zhang04} uses the sub-gradient computed at some data vector ${\bf x}_i$ to iteratively update ${\bf w}$. \citet{Shwartz11} proposed a stochastic sub-gradient descent SVM solver, Pegasos, that is reported to be among the fastest linear SVM solvers. {\em Cutting plane algorithms} \citep{Kelley60} solve the primal problem by successively tightening a piecewise linear approximation. It was employed by \citet{Joachims06} to solve linear SVMs with their implementation $\text{SVM}^{\text{perf}}$. This work was generalized in \citet{Joachims09} to include non-linear SVMs by approximately estimating $\bf w$ with arbitrary basis vectors using the fix-point iteration method \citep{Scholkopf01}. \citet{Teo10} proposed a related method for linear SVMs, that corrected some stability issues in the cutting plane methods.

\section{Analysis of AESVM}

As mentioned in the introduction, AESVM is an optimization problem on a subset of the training dataset called the representative set. In this section we first define the representative set. Then we present some properties of AESVM. These results are intended to provide theoretical justifications for the use of AESVM as an approximation to the SVM problem (\ref{eq:SVC_primal2}). We denote the cardinality of a set $S$ by $|S|$.

\subsection{Definition of the representative set} \label{sec:RPdef}

The convex hull of a set ${\bf X}$ is the smallest convex set containing ${\bf X}$ \citep{rockafellar} and can be obtained by taking all possible convex combinations of elements of ${\bf X}$. Assuming ${\bf X}$ is finite, the convex hull is a polygon. The extreme points of ${\bf X}$, $EP({\bf X})$, are defined to be the vertices of the convex polygon formed by the convex hull of ${\bf X}$. Any  vector ${\bf x}_i$ in ${\bf X}$ can be represented as a convex combination of vectors in $EP({\bf X})$: 
\begin{equation*}
{\bf x}_i = \underset{{\bf x}_t \in EP({\bf X})}{\sum} \pi^i _t {\bf x}_t \text{, where } 0 \le \pi^i _t \le 1 \text{, and } \underset{{\bf x}_t \in EP({\bf X})}{\sum} \pi^i _t = 1
\end{equation*}

We can see that functions of any data vector in ${\bf X}$ can be computed using only $EP({\bf X})$ and the convex combination weights $\{\pi^i _t\}$. The design of AESVM is motivated by the intuition that the use of extreme points may provide computational efficiency. However, extreme points are not useful in all cases, as for some kernels all data vectors are extreme points in kernel space. For example, for the Gaussian kernel, $K({\bf x}_i,{\bf x}_i) = \phi({\bf x}_i)^T\phi({\bf x}_i) = 1$. This implies that all the data vectors lie on the surface of the unit ball in the Gaussian kernel space and therefore are extreme points. Hence, we introduce the concept of {\em approximate extreme points}. 

Consider the set of transformed data vectors:
\begin{equation}\label{eq:Zdef}
{\bf Z} = \{{\bf z}_i:{\bf z}_i = \phi({\bf x}_i), \forall {\bf x}_i \in {\bf X}\}
\end{equation}
Here, the explicit representation of vectors in kernel space is only for the ease of understanding and all the computations are performed using kernel products. Let $V$ be a positive integer that is much smaller than $N$ and $\epsilon$ be a small positive real number. For notational simplicity, we assume $N$ is divisible by $V$.  Let ${\bf Z}_l$ be subsets of ${\bf Z}$ for $l = 1,2,...,(\frac{N}{V})$, such that ${\bf Z} = \underset{l}{\cup} {\bf Z}_l$ and ${\bf Z}_l \cap {\bf Z}_m = \emptyset$ for $l \neq m$, where $m = 1,2,...,(\frac{N}{V})$.  We require that the subsets ${\bf Z}_l$ satisfy  $|{\bf Z}_l| = V, \forall l$ and 
\begin{equation} \label{eq:RPyProp}
\forall {\bf z}_i, {\bf z}_j \in {\bf Z}_l,\text{ we have } y_i = y_j
\end{equation}  

Let ${\bf Z}_l ^q$ denote an arbitrary subset of ${\bf Z}_l$. Next, for any ${\bf z}_i \in {\bf Z}_l$ we define:
\begin{align} \label{eq:AEX}
&f({\bf z}_i,{\bf Z}_l ^q) = \underset{\overline{\mu^i}}{\textbf{min}}\|{\bf z}_i - \underset{{\bf z}_t \in {\bf Z}_l ^q}{\sum} \mu^i _t {\bf z}_t\|^2 \\
&\text{ s.t. } 0 \le \mu^i _t \le 1 \text{, and } \underset{{\bf z}_t \in {\bf Z}_l ^q}{\sum} \mu^i _t = 1 \nonumber
\end{align}
Consider the collection of subsets 
\begin{equation*}
{\cal Z}_\epsilon :=  \{{\bf Z}_l ^q: \underset{{\bf z}_i \in {\bf Z}_l}{\textbf{max }} f({\bf z}_i,{\bf Z}_l ^q) \le \epsilon\}
\end{equation*}
 A set of approximate extreme points of ${\bf Z}_l$ is denoted by ${\bf Z}_l ^*$, and is defined as follows \footnote{The properties derived for AESVM in Section \ref{sec:prop} are valid for any ${\bf Z}_l ^q$. The requirement for the smallest ${\bf Z}_l ^q$ is made only for the sake of a computationally simpler AESVM problem}
\begin{align} \label{eq:AEX2}
{\bf Z}_l ^* \in &\underset{{\bf Z}_l ^q \in {\cal Z}_\epsilon}{\textbf{ argmin }} |{\bf Z}_l ^q| 
%\text{over all  }\underset{{\bf z}_i \in {\bf Z}_l}{\textbf{max }} f({\bf z}_i,{\bf Z}_l ^q) \le \epsilon 
\end{align}
It can be seen that $\mu^i _t$ for ${\bf z}_t \in {\bf Z}_l ^*$ are  analogous to the convex combination weights $\pi^i _t$ for ${\bf x}_t \in EP({\bf X})$. The {\it representative set} ${\bf Z}^*$ of ${\bf Z}$ is the union of the sets of approximate extreme points of its subsets ${\bf Z}_l$.
\begin{equation*}\label{repZ}
{\bf Z}^* = \overset{\frac{N}{V}}{\underset{l = 1}{\cup}}{\bf Z}_l ^*
\end{equation*}

The representative set has properties that are similar to $EP({\bf X})$. Given any ${\bf z}_i \in {\bf Z}$, we can find ${\bf Z}_l$ such that ${\bf z}_i \in {\bf Z}_l$. Let $\gamma^i _t = \{\mu^i _t \text{ for } {\bf z}_t \in {\bf Z}_l ^* \text{ and }{\bf z}_i \in {\bf Z}_l \text{, and 0 otherwise}\}$. Now using (\ref{eq:AEX}), we can write:
\begin{equation}\label{eq:RKHS_AE}
{\bf z}_i = \underset{{\bf z}_t \in {\bf Z}^*}{\sum} \gamma^i _t{{\bf z}_t} + \tau_i
\end{equation}
Here $\tau_i$ is a vector that accounts for the approximation error $f({\bf z}_i,{\bf Z}_l ^q)$ in (\ref{eq:AEX}). From (\ref{eq:AEX})-(\ref{eq:RKHS_AE}) we can conclude that:
\begin{equation} \label{eq:tauNorm}
\|\tau_i\|^2 \le \epsilon \;\forall\; {\bf z}_i \in {\bf Z} 
\end{equation}
Since $\epsilon$ will be set to a very small positive constant, we can infer that $\tau_i$ is a very small vector.  The weights $\gamma^i _t$ are used to define $\beta_t$ in (\ref{eq:AESVM_primal2}) as:
\begin{equation} \label{eq:betaDef}
\beta_t = \overset{N}{\underset{i = 1}\sum} \gamma^i _t
\end{equation}

For ease of notation, we refer to the set ${\bf X}^* := \{{\bf x}_t:{\bf z}_t \in {\bf Z}^*\}$ as the representative set of ${\bf X}$ in the remainder of this paper. For the sake of simplicity, we assume that all $\gamma^i _t, \beta_t, {\bf X}, \text{ and } {\bf X}^*$ are arranged so that ${\bf X}^*$ is positioned as the first $M$ vectors of ${\bf X}$, where $M = |{\bf Z}^*|$.  

\subsection{Properties of AESVM}\label{sec:prop}
Consider the following optimization problem.
\begin{align}
&\underset{\mathbf{w} ,b}{\mathbf{min}}\;F_3 (\mathbf{w}, b) = \frac{1}{2}{\|{\mathbf{w}}\|}^{2}\ + \frac{C}{N} \overset{N}{\underset{i = 1}\sum} l(\mathbf{w},b,{\bf u}_i) \label{eq:F3_primal}\\
&\text{where }{\bf u}_i = \overset{M}{\underset{t = 1}\sum} \gamma^i _t  {\bf z}_t, {\bf z}_t \in {\bf Z}^*,{\bf w} \in \mathbb{H}\text{, and } b \in \mathbb{R}\nonumber
\end{align}
We use the problem in (\ref{eq:F3_primal}) as an intermediary between (\ref{eq:SVC_primal2}) and (\ref{eq:AESVM_primal2}). The intermediate problem (\ref{eq:F3_primal}) has a direct relation to the AESVM problem, as given in the following theorem. The properties of the $max$ function given below are relevant to the following discussion: 
\begin{align}
&max(0,A+B) \le max(0,A) + max(0,B) \label{eq:maxA}\\ 
&max(0,A-B) \ge max(0,A) - max(0,B) \label{eq:maxB}\\
&\overset{N}{\underset{i = 1}{\sum}} max(0,c^i A) = max(0,A) \overset{N}{\underset{i = 1}{\sum}}c^i \label{eq:maxC}
\end{align}
for $A,B,c^i \in \mathbb{R}$ and $c^i \ge 0$. 

\noindent
{\bf Theorem 1} {\it Let $F_3 (\mathbf{w}, b)$ and $F_2 (\mathbf{w}, b)$ be as defined in (\ref{eq:F3_primal}) and (\ref{eq:AESVM_primal2}) respectively. Then, 
\begin{equation*}
F_3 (\mathbf{w}, b) \le F_2 (\mathbf{w}, b)\;, \forall {\bf w} \in \mathbb{H} \text{ and } b \in \mathbb{R}
\end{equation*}}

\begin{proof}
Let ${\cal L}_2({\bf w},b,{\bf X}^*) = \frac{C}{N}\overset{M}{\underset{t = 1}\sum}l(\mathbf{w},b,{\bf z}_t)\overset{N}{\underset{i = 1}\sum} \gamma^i _t$ and ${\cal L}_3({\bf w},b,{\bf X}^*) = \frac{C}{N}\overset{N}{\underset{i = 1}\sum}l(\mathbf{w},b,{\bf u}_i)$, where ${\bf u}_i = \overset{M}{\underset{t = 1}\sum} \gamma^i _t {\bf z}_t$. From the properties of $\gamma^i _t$ in (\ref{eq:AEX}), and from (\ref{eq:RPyProp}) we get:
\begin{align}
{\cal L}_3({\bf w},b,{\bf X}^*) &= \frac{C}{N}\overset{N}{\underset{i = 1}{\sum}}  max\left[0, \left\{1- y_{i}(\mathbf{w}^T { \overset{M}{\underset{t = 1}\sum} \gamma^i _t {\bf z}_t} + b)\right\}\right] \label{eq:L3_def}\\
&=\frac{C}{N} \overset{N}{\underset{i = 1}{\sum}} max\left[0, \overset{M}{\underset{t = 1}\sum} \gamma^i _t\left\{1- y_{t}(\mathbf{w}^T {{\bf z}_t} + b)\right\}\right]\nonumber
\end{align}

Using properties (\ref{eq:maxA}) and (\ref{eq:maxC}) we get:
\begin{align}
{\cal L}_3({\bf w},b,{\bf X}^*) &\le \frac{C}{N}\overset{N}{\underset{i = 1}{\sum}} \overset{M}{\underset{t = 1}\sum} max\left[0, \gamma^i _t \left\{1- y_{t}(\mathbf{w}^T {{\bf z}_t} + b)\right\}\right] \nonumber\\
&= \frac{C}{N} \overset{M}{\underset{t = 1}\sum} max\left[0, 1- y_{t}(\mathbf{w}^T {{\bf z}_t} + b)\right] \overset{N}{\underset{i = 1}{\sum}}\gamma^i _t \nonumber\\
&= {\cal L}_2({\bf w},b,{\bf X}^*) \nonumber
\end{align}
Adding $\frac{1}{2}{\|{\mathbf{w}}\|}^{2}$ to both sides of the inequality above we get
\begin{equation*}
F_3 (\mathbf{w}, b) \le F_2 (\mathbf{w}, b)
\end{equation*}
\end{proof}

\noindent
The following theorem gives a relationship between the SVM problem and the intermediate problem.

\noindent
{\bf Theorem 2} {\it Let $F_1 (\mathbf{w}, b)$ and $F_3 (\mathbf{w}, b)$ be as defined in (\ref{eq:SVC_primal2}) and (\ref{eq:F3_primal}) respectively. Then, 
\begin{align*}
&-\frac{C}{N}\overset{N}{\underset{i = 1}{\sum}}max\left\{0, y_{i}\mathbf{w}^T \tau_i\right\} \le F_1 (\mathbf{w}, b) - F_3 (\mathbf{w}, b) \le \frac{C}{N}\overset{N}{\underset{i = 1}{\sum}}max\left\{0, -y_{i}\mathbf{w}^T \tau_i\right\}\\
&\forall {\bf w} \in \mathbb{H} \text{ and } b \in \mathbb{R}, \text{ where } \tau_i \in \mathbb{H} \text{ is the vector defined in (\ref{eq:RKHS_AE}).}
\end{align*}
}

\begin{proof}
Let ${\cal L}_1({\bf w},b,{\bf X}) = \frac{C}{N}\overset{N}{\underset{i = 1}\sum}l(\mathbf{w},b,{\bf z}_i)$, denote the average hinge loss that is minimized in (\ref{eq:SVC_primal2}) and ${\cal L}_3({\bf w},b,{\bf X}^*)$ be as defined in Theorem 1. Using (\ref{eq:RKHS_AE}) and (\ref{eq:SVC_primal2}) we get:
\begin{align*}
{\cal L}_1({\bf w},b,{\bf X}) &= \frac{C}{N}\overset{N}{\underset{i = 1}\sum} max\left\{0, 1- y_{i}(\mathbf{w}^T \mathbf{z_{i}} + b)\right\}\\
									&= \frac{C}{N}\overset{N}{\underset{i = 1}\sum} max\left\{0, 1- y_{i}(\mathbf{w}^T (\overset{M}{\underset{t = 1}{\sum}} \gamma^i _t{{\bf z}_t} + \tau_i) + b)\right\}
\end{align*}									
From the properties of $\gamma^i _t$ in (\ref{eq:AEX}), and from (\ref{eq:RPyProp}) we get:
\begin{equation}
{\cal L}_1({\bf w},b,{\bf X}) = \frac{C}{N}\overset{N}{\underset{i = 1}\sum} max\left\{0, \overset{M}{\underset{t = 1}{\sum}} \gamma^i _t (1- y_{t}(\mathbf{w}^T {{\bf z}_t} + b)) - y_{i}\mathbf{w}^T \tau_i\right\} \label{eq:T1E1}
\end{equation}
Using (\ref{eq:maxA}) on (\ref{eq:T1E1}), we get:
\begin{align*}
{\cal L}_1({\bf w},b,{\bf X}) &\le \frac{C}{N}\overset{N}{\underset{i = 1}{\sum}}  max\left[0, \overset{M}{\underset{t = 1}\sum} \gamma^i _t\left\{1- y_{t}(\mathbf{w}^T {{\bf z}_t} + b)\right\}\right] + \frac{C}{N}\overset{N}{\underset{i = 1}{\sum}}max\left\{0, -y_{i}\mathbf{w}^T \tau_i\right\}\\
&= {\cal L}_3({\bf w},b,{\bf X}^*) + \frac{C}{N}\overset{N}{\underset{i = 1}{\sum}}max\left\{0, -y_{i}\mathbf{w}^T \tau_i\right\}
\end{align*}
Using (\ref{eq:maxB}) on (\ref{eq:T1E1}), we get:
\begin{align*}
{\cal L}_1({\bf w},b,{\bf X}) &\ge \frac{C}{N}\overset{N}{\underset{i = 1}{\sum}}  max\left[0, \overset{M}{\underset{t = 1}\sum} \gamma^i _t\left\{1- y_{t}(\mathbf{w}^T {{\bf z}_t} + b)\right\}\right] - \frac{C}{N}\overset{N}{\underset{i = 1}{\sum}}max\left\{0, y_{i}\mathbf{w}^T \tau_i\right\}\\
&= {\cal L}_3({\bf w},b,{\bf X}^*) - \frac{C}{N}\overset{N}{\underset{i = 1}{\sum}}max\left\{0, y_{i}\mathbf{w}^T \tau_i\right\}
\end{align*}
From the two inequalities above we get,
\begin{equation*}
{\cal L}_3({\bf w},b,{\bf X}^*) - \frac{C}{N}\overset{N}{\underset{i = 1}{\sum}}max\left\{0, y_{i}\mathbf{w}^T \tau_i\right\} \le {\cal L}_1({\bf w},b,{\bf X}) \le {\cal L}_3({\bf w},b,{\bf X}^*) + \frac{C}{N}\overset{N}{\underset{i = 1}{\sum}}max\left\{0, -y_{i}\mathbf{w}^T \tau_i\right\}
\end{equation*}
Adding $\frac{1}{2}{\|{\mathbf{w}}\|}^{2}$ to the inequality above we get
\begin{equation*}
F_3 (\mathbf{w}, b)- \frac{C}{N}\overset{N}{\underset{i = 1}{\sum}}max\left\{0, y_{i}\mathbf{w}^T \tau_i\right\} \le F_1 (\mathbf{w}, b) \le F_3 (\mathbf{w}, b) + \frac{C}{N}\overset{N}{\underset{i = 1}{\sum}}max\left\{0, -y_{i}\mathbf{w}^T \tau_i\right\}
\end{equation*}
\end{proof}

Using the above theorems we derive the following corollaries. These results provide the theoretical justification for AESVM.

\noindent
{\bf Corollary 3} {\it Let $({\bf w}_1 ^*, b _1 ^*)$ be the solution of (\ref{eq:SVC_primal2}) and   $({\bf w}_2 ^*, b _2 ^*)$ be the solution of (\ref{eq:AESVM_primal2}). Then, 
\begin{equation*}
F_1 ({\bf w}_1 ^*, b _1 ^*) - F_2 ({\bf w}_2 ^*, b _2 ^*) \le C\sqrt{C\epsilon}
\end{equation*}}

\begin{proof}
It is known that $\|{\bf w}_1^*\| \le \sqrt{C}\;$ (refer Theorem 1 in \citet{Shwartz11}). It is straight forward to see that the same result also applies to AESVM, $\|{\bf w}_2^*\| \le \sqrt{C}\;$. Based on (\ref{eq:tauNorm}) we know that $\|\tau_i\| \le \sqrt{\epsilon}$. From Theorem 2 we get: 
\begin{align}
F_1 ({\bf w}_2 ^*, b _2 ^*) - F_3 ({\bf w}_2 ^*, b _2 ^*) &\le \frac{C}{N}\overset{N}{\underset{i = 1}{\sum}}max\left\{0, -y_{i}{\bf w}_2 ^{*T} \tau_i\right\} \le \frac{C}{N}\overset{N}{\underset{i = 1}{\sum}} \|{\bf w}_2^*\|\|\tau_i\| \nonumber\\
&\le \frac{C}{N}\overset{N}{\underset{i = 1}{\sum}} \sqrt{C\epsilon} = C\sqrt{C\epsilon} \nonumber
\end{align}
Since $({\bf w}_1 ^*, b _1 ^*)$ is the solution of (\ref{eq:SVC_primal2}), $F_1 ({\bf w}_1 ^*, b _1 ^*) \le F_1 ({\bf w}_2 ^*, b _2 ^*)$. Using this property and Theorem 1 in the inequality above, we get:
\begin{align}
F_1 ({\bf w}_1 ^*, b _1 ^*) - F_2 ({\bf w}_2 ^*, b _2 ^*) &\le F_1 ({\bf w}_1 ^*, b _1 ^*) - F_3 ({\bf w}_2 ^*, b _2 ^*) \nonumber\\
&\le F_1 ({\bf w}_2 ^*, b _2 ^*) - F_3 ({\bf w}_2 ^*, b _2 ^*) \le C\sqrt{C\epsilon} \label{eq:cor2.1}
\end{align}
\end{proof}

Now we demonstrate some properties of AESVM using the dual problem formulations of AESVM and the intermediate problem. The dual form of AESVM is given by:
\begin{align} \label{eq:AESVM_dual}
\underset{\bf{\hat{\alpha}}}{\textbf{max }} L_2(\hat{\alpha}) =  \overset{M}{\underset{t = 1}{\sum}}\hat{\alpha}_{t} - \frac{1}{2}\overset{M}{\underset{t = 1}{\sum}}\overset{M}{\underset{s = 1}{\sum}}\hat{\alpha}_{t}\hat{\alpha}_{s}y_{t}y_{s} {\bf z}^T _t {\bf z}_s\\
\text{ subject to } 0 \le \hat{\alpha}_t \le \frac{C}{N}\overset{N}{\underset{i = 1}\sum} \gamma^i _t \text{ and } \underset{t = 1}{\overset{M}{\sum}}\hat{\alpha}_{t}y_t = 0  \nonumber
\end{align}
The dual form of the intermediate problem is given by:
\begin{align} \label{eq:F3_dual}
&\underset{\breve{\alpha}}{\textbf{max }} L_3(\breve{\alpha}) =  \overset{N}{\underset{i = 1}{\sum}}\breve{\alpha}_{i} - \frac{1}{2}\overset{N}{\underset{i = 1}{\sum}}\overset{N}{\underset{j = 1}{\sum}}\breve{\alpha}_{i}\breve{\alpha}_{j}y_{i}y_{j}{\bf u}^T _i {\bf u}_j\\
&\text{ subject to } 0 \le \breve{\alpha}_i \le \frac{C}{N} \text{ and } \underset{i = 1}{\overset{N}{\sum}}\breve{\alpha}_{i}y_i = 0  \nonumber
\end{align}
Consider the mapping function $h: \mathbb{R}^N \rightarrow \mathbb{R}^M$, defined as 
\begin{equation}\label{eq:hfunc}
h(\breve{\alpha}) = \{\tilde{\alpha}_t: \tilde{\alpha}_t = \overset{N}{\underset{i = 1}\sum} \gamma^i _t \breve{\alpha}_i\}
\end{equation}
It can be seen that the objective functions $L_2(h(\breve{\alpha}))$ and $L_3(\breve{\alpha})$ are identical.
\begin{align*}
L_2(h(\breve{\alpha})) &= \overset{M}{\underset{t = 1}{\sum}}\tilde{\alpha}_{t} - \frac{1}{2}\overset{M}{\underset{t = 1}{\sum}}\overset{M}{\underset{s = 1}{\sum}}\tilde{\alpha}_{t}\tilde{\alpha}_{s}y_{t}y_{s} {\bf z}^T _t {\bf z}_s\\
&= \overset{N}{\underset{i = 1}{\sum}}\breve{\alpha}_{i} - \frac{1}{2}\overset{N}{\underset{i = 1}{\sum}}\overset{N}{\underset{j = 1}{\sum}}\breve{\alpha}_{i}\breve{\alpha}_{j}y_{i}y_{j} {\bf u}^T _i {\bf u}_j\\
&= L_3(\breve{\alpha})
\end{align*}
It is also straight forward to see that, for any feasible $\breve{\alpha}$ of (\ref{eq:F3_dual}), $h(\breve{\alpha})$ is a feasible point of (\ref{eq:AESVM_dual}) as it satisfies the constraints in (\ref{eq:AESVM_dual}). However, the converse is not always true. With that clarification, we present the following corollary.

\noindent
{\bf Corollary 4} {\it Let $({\bf w}_1 ^*, b _1 ^*)$ be the solution of (\ref{eq:SVC_primal2}) and $({\bf w}_2 ^*, b _2 ^*)$ be the solution of (\ref{eq:AESVM_primal2}). Let $\hat{\alpha}_2$ be the dual variable corresponding to $({\bf w}_2 ^*, b _2 ^*)$. Let $h(\breve{\alpha}_2)$ be as defined in (\ref{eq:hfunc}). If there exists an $\breve{\alpha}_2$ such that $h(\breve{\alpha}_2) = \hat{\alpha}_2$ and $\breve{\alpha}_2$ is a feasible point of (\ref{eq:F3_dual}), then, 
\begin{equation*}
F_1 ({\bf w}_2 ^*, b _2 ^*) - F_1 ({\bf w}_1 ^*, b _1 ^*) \le 2C\sqrt{C\epsilon}
\end{equation*}}

\begin{proof}
Let $({\bf w}_3 ^*, b _3 ^*)$ be the solution of (\ref{eq:F3_primal}) and $\breve{\alpha}_3$ the solution of (\ref{eq:F3_dual}). We know that $L_3(\breve{\alpha}_2) = L_2(\hat{\alpha}_2) = F_2 ({\bf w}_2 ^*, b _2 ^*)$ and $L_3(\breve{\alpha}_3) = F_3 ({\bf w}_3 ^*, b _3 ^*)$. Since $L_3(\breve{\alpha}_3) \ge L_3(\breve{\alpha}_2)$, we get 
\begin{equation*}
F_3 ({\bf w}_3 ^*, b _3 ^*) \ge F_2 ({\bf w}_2 ^*, b _2 ^*) 
\end{equation*}
But, from Theorem 1 we know $F_3 ({\bf w}_3 ^*, b _3 ^*) \le F_3 ({\bf w}_2 ^*, b _2 ^*) \le F_2 ({\bf w}_2 ^*, b _2 ^*)$. Hence 
\begin{equation*}
F_3 ({\bf w}_3 ^*, b _3 ^*) = F_3 ({\bf w}_2 ^*, b _2 ^*)
\end{equation*}
From the above result we get 
\begin{equation} \label{eq:cor4e1}
F_3 ({\bf w}_2 ^*, b _2 ^*) - F_3 ({\bf w}_1 ^*, b _1 ^*) \le 0 
\end{equation}
From Theorem 2 we have the following inequalities: 
\begin{align}
- \frac{C}{N}\overset{N}{\underset{i = 1}{\sum}}max\left\{0, y_{i}\mathbf{w}_1 ^{*T} \tau_i\right\} &\le F_1 ({\bf w}_1 ^*, b _1 ^*) - F_3 ({\bf w}_1 ^*, b _1 ^*) \label{eq:cor2e1}\\
F_1 ({\bf w}_2 ^*, b _2 ^*) - F_3 ({\bf w}_2 ^*, b _2 ^*) &\le \frac{C}{N}\overset{N}{\underset{i = 1}{\sum}}max\left\{0, -y_{i}\mathbf{w}_2 ^{*T} \tau_i\right\} \label{eq:cor2e2}
\end{align}
Adding (\ref{eq:cor2e1}) and (\ref{eq:cor2e2}) we get:
\begin{align} 
F_1 ({\bf w}_2 ^*, b _2 ^*) - F_1 ({\bf w}_1 ^*, b _1 ^*) &\le R + \frac{C}{N}\overset{N}{\underset{i = 1}{\sum}} \left[ max\left\{0, -y_{i}\mathbf{w}_2 ^{*T} \tau_i\right\} + max\left\{0, y_{i}\mathbf{w}_1 ^{*T} \tau_i\right\} \right]\label{eq:cor2}
\end{align}
where $R = F_3 ({\bf w}_2 ^*, b _2 ^*) - F_3 ({\bf w}_1 ^*, b _1 ^*)$. Using (\ref{eq:cor4e1}) and the properties
 $\|{\bf w}_2 ^*\| \le \sqrt{C}$ and $\|{\bf w}_1 ^*\| \le \sqrt{C}\;$ in (\ref{eq:cor2}): 
\begin{align*}
F_1 ({\bf w}_2 ^*, b _2 ^*) - F_1 ({\bf w}_1 ^*, b _1 ^*) &\le \frac{C}{N}\overset{N}{\underset{i = 1}{\sum}} \left[ max\left\{0, -y_{i}{\bf w}_2 ^{*T} \tau_i\right\} + max\left\{0, y_{i}{\bf w}_1 ^{*T} \tau_i\right\}\right]\\
&\le \frac{C}{N}\overset{N}{\underset{i = 1}{\sum}} \|{\bf w}_2^*\| \|\tau_i\| + \|{\bf w}_1^*\| \|\tau_i\|\\
&\le \frac{C}{N}\overset{N}{\underset{i = 1}{\sum}} 2\sqrt{C\epsilon} = 2C\sqrt{C\epsilon}
\end{align*}
\end{proof}

Now we prove a relationship between AESVM and the Gram matrix approximation methods mentioned in Section \ref{sec:otherDual}. 

\noindent
{\bf Corollary 5} {\it Let $L_1(\alpha)$, $L_3(\breve{\alpha})$, and ${F}_2(\mathbf{w} ,b)$ be the objective functions of the SVM dual (\ref{eq:SVC_dual}), intermediate dual (\ref{eq:F3_dual}) and AESVM (\ref{eq:AESVM_primal2}) respectively. Let ${\bf z}_i$, $\tau_i$, and ${\bf u}_i$ be as defined in (\ref{eq:Zdef}), (\ref{eq:RKHS_AE}), and (\ref{eq:F3_primal}) respectively. Let $\bf G$ and $\tilde{\bf G}$ be the $N\mathrm{x}N$ matrices with ${\bf G}_{ij} = {\bf y}_i {\bf y}_j {\bf z}_i ^T {\bf z}_j$ and $\tilde{\bf G}_{ij} = {\bf y}_i {\bf y}_j {\bf u}_i ^T {\bf u}_j$ respectively. Then for any feasible $\breve{\alpha},\alpha, \mathbf{w}, \text{and } b$:
\begin{enumerate}
\item Rank of $\tilde{\bf G} = M, L_1(\alpha) = \overset{N}{\underset{i = 1}{\sum}}\alpha_{i} - \frac{1}{2}\alpha {\bf G} \alpha ^T, L_3(\breve{\alpha}) = \overset{N}{\underset{i = 1}{\sum}}\breve{\alpha}_{i} - \frac{1}{2}\breve{\alpha} \tilde{\bf G} \breve{\alpha}^T$, and
\begin{equation*}
\mathrm{Trace}({\bf G} - \tilde{\bf G}) \le N\epsilon + 2 \overset{M}{\underset{t = 1}\sum} {\bf z}_t ^T \overset{N}{\underset{i = 1}{\sum}} \gamma^i _t \tau_i
\end{equation*}
\item ${F}_2(\mathbf{w} ,b) \ge L_3(\breve{\alpha})$
\end{enumerate}
}

\begin{proof}
Using $\bf G$, the SVM dual objective function $L_1(\alpha)$ can be represented as:
\begin{equation*}
L_1(\alpha) = \overset{N}{\underset{i = 1}{\sum}}\alpha_{i} - \frac{1}{2}\alpha {\bf G} \alpha ^T
\end{equation*}
Similarly, $L_3(\breve{\alpha})$ can be represented using $\tilde{\bf G}$ as: 
\begin{equation*}
L_3(\breve{\alpha}) = \overset{N}{\underset{i = 1}{\sum}}\breve{\alpha}_{i} - \frac{1}{2}\breve{\alpha} \tilde{\bf G} \breve{\alpha}^T
\end{equation*}
Applying ${\bf u}_i = \overset{M}{\underset{t = 1}\sum} \gamma^i _t {\bf z}_t, \;\forall {\bf z}_t \in {\bf Z}^*$ to the definition of $\tilde{\bf G}$, we get:
\begin{equation*}
\tilde{\bf G} = \Gamma {\bf A} \Gamma ^T
\end{equation*}
Here ${\bf A}$ is the $M$x$M$ matrix comprised of ${\bf A}_{ts} = {\bf y}_t {\bf y}_s {\bf z}_t ^T {\bf z}_s, \; \forall {\bf z}_t, {\bf z}_s \in {\bf Z}^*$ and $\Gamma$ is the $N$x$M$ matrix with the elements $\Gamma_{it} = \gamma_t ^i$. Hence the rank of $\tilde{\bf G} = M$ and intermediate dual problem (\ref{eq:F3_dual}) is a low rank approximation of the SVM dual problem (\ref{eq:SVC_dual}). 

The Gram matrix approximation error can be quantified using (\ref{eq:RKHS_AE}) and (\ref{eq:tauNorm}) as:
\begin{align*}
\mathrm{Trace}({\bf G} - \tilde{\bf G}) &= \overset{N}{\underset{i = 1}{\sum}} \left[{\bf z}_i ^T {\bf z}_i - (\overset{M}{\underset{t = 1}\sum} \gamma^i _t {\bf z}_t) ^T (\overset{M}{\underset{s = 1}\sum} \gamma^i _s{\bf z}_s)\right]\\
%\gamma^i _s{\bf z}_s + \tau_i)- (\overset{M}{\underset{t = 1}\sum} \gamma^i _t {\bf z}_t) ^T (\overset{M}{\underset{s = 1}\sum} \gamma^i _s{\bf z}_s)\right]\\ 
& = \overset{N}{\underset{i = 1}{\sum}} \left[\tau_i ^T \tau_i + 2\overset{M}{\underset{t = 1}\sum} \gamma^i _t {\bf z}_t ^T \tau_i\right] \le N\epsilon + 2 \overset{M}{\underset{t = 1}\sum} {\bf z}_t ^T \overset{N}{\underset{i = 1}{\sum}} \gamma^i _t \tau_i
\end{align*}

By the principle of duality, we know that ${F}_3(\mathbf{w} ,b) \ge L_3(\breve{\alpha}), \; \forall {\bf w} \in \mathbb{H} \text{ and } b \in \mathbb{R}$, where $\breve{\alpha}$ is any feasible point of (\ref{eq:F3_dual}). Using Theorem 1 on the inequality above, we get
\begin{equation*}
{F}_2(\mathbf{w} ,b) \ge L_3(\breve{\alpha}), \; \forall {\bf w} \in \mathbb{H}, b \in \mathbb{R} \text{ and feasible $\breve{\alpha}$} 
\end{equation*}
Thus the AESVM problem minimizes an upper bound (${F}_2(\mathbf{w} ,b)$) of a rank $M$ Gram matrix approximation of $L_1(\alpha)$.
\end{proof}

Based on the theoretical results in this section, it is reasonable to suggest that for small values of $\epsilon$, the solution of AESVM is close to the solution of SVM.

\section{Computation of the representative set} \label{sec:derRP}
In this section, we present algorithms to compute the representative set. {\em The AESVM formulation can be solved with any standard SVM solver such as SMO and hence we do not discuss methods to solve it}. As described in Section \ref{sec:RPdef}, we require an algorithm to compute approximate extreme points in kernel space. \citet{OSUNA02} proposed an algorithm to derive extreme points of the convex hull of a dataset in kernel space. Their algorithm is computationally intensive, with a time complexity of $O(N\;S(N))$, and is unsuitable for large datasets as $S(N)$ typically has a super-linear dependence on N. The function $S(N)$ denotes the time complexity of a SVM solver (required by their algorithm), to train on a dataset of size N. We next propose two algorithms leveraging the work by \citet{OSUNA02} to compute the representative set in kernel space ${\bf Z}^*$ with much smaller time complexities.

We followed the divide and conquer approach to develop our algorithms. The dataset is first divided into subsets ${\bf X}_q, q = 1,2,..,Q$, where $|{\bf X}_q| < P$, $Q \ge \frac{N}{P}$ and ${\bf X} = \{ {\bf X}_1, {\bf X}_2, .., {\bf X}_Q\}$. The parameter $P$ is a predefined large integer. It is desired that each subset ${\bf X}_q$ contains data vectors that are more similar to each other than data vectors in other subsets. Our notion of similarity of data vectors in a subset, is that the distances between data vectors within a subset is less than the distances between data vectors in distinct subsets. This first level of segregation is followed by another level of segregation. We can regard the first level of segregation as coarse segregation and the second as fine segregation. Finally, the approximate extreme points of the subsets obtained after segregation, are computed. The two different algorithms to compute the representative set differ only in the first level of segregation as described in the following sections.

\subsection{First level of segregation}
We propose the methods, FLS1 and FLS2 given below to perform a first level of segregation. In the following description we use arrays $\Delta'$ and $\Delta'_2$ of $N$ elements. Each element of $\Delta'$ ($\Delta' _2$), $\delta_i$ ($\delta_i ^2$) , contains the index in ${\bf X}$ of the last data vector of the subset to which ${\bf x}_i$ belongs. It is straight forward to replace this $N$ element array with a smaller array of size equal to the number of subsets. We use a $N$ element array for ease of description.\newline

\noindent
{\it 1. FLS1(${\bf X}' ,P$)}

For some applications, such as anomaly detection on sequential data, data vectors are found to be homogeneous within intervals. For example, the atmospheric conditions typically do not change within a few minutes and hence weather data is homogeneous for a short span. For such datasets it is enough to segregate the data vectors based on its position in the training dataset. The same method can also be used on very large datasets without any homogeneity, in order to reduce computation time. The complexity of this method is $O(N' )$, where $N'  = |{\bf X}' |$ . 
\floatname{algorithm}{ }
\begin{algorithm}[h]
 \renewcommand{\thealgorithm}{}
\caption{[${\bf X}' $,$\Delta' $] = FLS1(${\bf X}' ,P$)}
\begin{enumerate} \addtolength{\itemsep}{-.35\baselineskip} 
  \item For outerIndex = 1 {\bf{\emph to}} ceiling($\frac{|{\bf X}' |}{P}$)
  \item \hspace{5pt} For innerIndex = (outerIndex - 1)$P$ {\bf{\emph to}} min((outerIndex)$P$,$|{\bf X}' |$)  
  \item \hspace{15pt} Set $\delta_{innerIndex} = min((outerIndex)P,|{\bf X}' |)$
\end{enumerate}
\end{algorithm}

\noindent
{\it 2. FLS2(${\bf X}' ,P$)}

When the dataset is not homogeneous within intervals or it is not excessively large we use the more sophisticated algorithm, FLS2, of time complexity $O(N' \; \text{log}_2 \frac{N' }{P})$ given below. In step 1 of FLS2, the distance $d_i$ in kernel space of all ${\bf x}_i \in {\bf X}' $ from ${\bf x}_j$ is computed as $d_i = \|\phi({\bf x}_i) - \phi({\bf x}_j)\|^2 = k({\bf x}_i, {\bf x}_i) + k({\bf x}_j, {\bf x}_j) - 2k({\bf x}_i, {\bf x}_j)$. The algorithm FLS2(${\bf X}' ,P$), in effect builds a binary search tree, with each node containing the data vector ${\bf x}_k$ selected in step 2 that partitions a subset of the dataset into two. The size of the subsets successively halve, on downward traversal from the root of the tree to the other nodes. When the size of all the subsets at a level become $\le P$ the algorithm halts. The complexity of FLS2 can be derived easily when the algorithm is considered as an incomplete binary search tree building method. The last level of such a tree will have $O(\frac{N' }{P})$ nodes and consequently the height of the tree is $O(\text{log}_2 \frac{N' }{P})$. At each level of the tree the calls to the BFPRT algorithm \citep{Blum73} and the rearrangement of the data vectors in steps 2 and 3 are of $O(N' )$ time complexity. Hence the overall time complexity of FLS2(${\bf X}' ,P$) is $O(N' \; \text{log}_2 \frac{N' }{P})$. \newline

\floatname{algorithm}{ }
\begin{algorithm}[h]
 \renewcommand{\thealgorithm}{}
\caption{[${\bf X}' $,$\Delta' $] = FLS2(${\bf X}' ,P$)}
\begin{enumerate} \addtolength{\itemsep}{-.35\baselineskip} 
 \item Compute distance $d_i$ in kernel space of all ${\bf x}_i \in {\bf X}' $ from the first vector ${\bf x}_j$ in ${\bf X}' $
 \item Select ${\bf x}_k$ such that there exists $\frac{|{\bf X}' |}{2}$ data vectors ${\bf x}_i \in {\bf X}' $ with $d_i < d_k$, using the linear time BFPRT algorithm
 \item Using ${\bf x}_k$, rearrange ${\bf X}' $ as ${\bf X}'  = \{{\bf X}^1,{\bf X}^2\}$, where ${\bf X}^1 = \{{\bf x}_i: d_i < d_k, {\bf x}_i \in {\bf X}' \}$ and ${\bf X}^2 = \{{\bf x}_i: {\bf x}_i \in {\bf X}'  \text{ and } {\bf x}_i \not\in {\bf X}^1\}$
 \item If $\frac{|{\bf X}' |}{2} \le P$
 \subitem For $i$ where ${\bf x}_i \in {\bf X}^1$, set $\delta_i$ = index of last data vector in ${\bf X}^1$.
 \subitem For $i$ where ${\bf x}_i \in {\bf X}^2$, set $\delta_i$ = index of last data vector in ${\bf X}^2$.
 \item If $\frac{|{\bf X}' |}{2} > P$
 \subitem Run FLS2(${\bf X}^1,P$) and FLS2(${\bf X}^2,P$)
\end{enumerate}
\end{algorithm}

\subsection{Second level of segregation}
After the initial segregation, another method SLS(${\bf X}' ,V,\Delta' $) is used to further segregate each set ${\bf X}_q$ into smaller subsets ${\bf X}_{q_r}$ of maximum size $V$, ${\bf X}_q = \{{\bf X}_{q_1}, {\bf X}_{q_2},....,{\bf X}_{q_R}\}$, where $V$ is predefined ($V < P$) and $R = ceiling(\frac{|{\bf X_q}|}{V})$. The algorithm SLS(${\bf X}' ,V,\Delta' $) is given below. In step 2.b, ${\bf x}_t$ is the data vector in ${\bf X}_q$ that is farthest from the origin in the space of the data vectors. For some kernels, such as the Gaussian kernel, all data vectors are equidistant from the origin in kernel space. If the algorithm chooses ${\bf a}^l$ in step 2.b based on distances in such kernel spaces, the choice would be arbitrary and such a situation is avoided here. Each iteration of the For loop in step 2 involves several runs of the BFPRT algorithm, with each run followed by a rearrangement of ${\bf X}_q$. Specifically, the BFPRT algorithm is first run on $P$ data vectors, then on $P - V$ data vectors, then on $P - 2V$ data vectors and so on. The time complexity of each iteration of the For loop including the BFPRT algorithm run and the rearrangement of data vectors is: $O(P + (P - V) + (P- 2V) + ..+ V) \Rightarrow O(\frac{P^2}{V})$. The overall complexity of SLS(${\bf X}' ,V,\Delta' $) considering the Q For loop iterations is $O(\frac{N' }{P}\frac{P^2}{V})  \Rightarrow O(\frac{N' P}{V})$, since $Q = O(\frac{N' }{P})$. \newline

\floatname{algorithm}{ }
\begin{algorithm}[h]
 \renewcommand{\thealgorithm}{}
\caption{[${\bf X}' $,$\Delta' _2$] = SLS(${\bf X}' ,V,\Delta' $)}
\begin{enumerate} \addtolength{\itemsep}{-.35\baselineskip}
 \item Initialize $l = 1$
 \item For q = 1 {\bf{\emph to}} $Q$
 \begin{enumerate}
 \item Identify subset ${\bf X}_q$ of ${\bf X}' $ using $\Delta' $ 
 \item Set ${\bf a}^l = \phi({\bf x}_t)$, where ${\bf x}_t \in \underset{i}{\textbf{argmax }}\|{\bf x}_i \|^2, {\bf x}_i \in {\bf X}_q$
 \item Compute distance $d_i$ in kernel space of all ${\bf x}_i \in {\bf X}_q$ from ${\bf a}^l$
 \item Select ${\bf x}_k$ such that, there exists $V$ data vectors ${\bf x}_i \in {\bf X}_q$ with $d_i < d_k$, using the BFPRT algorithm  
 \item Using ${\bf x}_k$, rearrange ${\bf X}_q$ as ${\bf X}_q = \{{\bf X}^1,{\bf X}^2\}$, where ${\bf X}^1 = \{{\bf x}_i: d_i < d_k, {\bf x}_i \in {\bf X}_q\}$ and ${\bf X}^2 = \{{\bf x}_i: {\bf x}_i \in {\bf X}_q \text{ and } {\bf x}_i \not\in {\bf X}^1\}$
 \item For $i$ where ${\bf x}_i \in {\bf X}^1$, set $\delta_i ^2$ = index of last data vector in ${\bf X}^1$, where $\delta_i ^2$ is the $i^{th}$ element of $\Delta' _2$
 \item Remove  ${\bf X}^1$ from ${\bf X}_q$
 \item If $|{\bf X}^2| > V$
 \subitem Set: $l = l + 1$ and ${\bf a}^l = {\bf x}_k$
 \subitem Repeat steps 2.c to 2.h
 \item If $|{\bf X}^2| \le V$
\subitem For $i$ where ${\bf x}_i \in {\bf X}^2$, set $\delta_i ^2$ = index of last data vector in ${\bf X}^2$
 \end{enumerate} 
\end{enumerate}
\end{algorithm}

\subsection{Computation of the approximate extreme points}

After computing the subsets ${\bf X}_{q_r}$, the algorithm DeriveAE is applied to each ${\bf X}_{q_r}$ to compute its approximate extreme points. The algorithm DeriveAE is described below. DeriveAE uses three routines. SphereSet(${\bf X}_{q_r}$) returns all ${\bf x}_i \in {\bf X}_{q_r}$ that lie on the surface of the smallest hypersphere in kernel space that contains ${\bf X}_{q_r}$. It computes the hypersphere as a hard margin support vector data descriptor (SVDD) \citep{Tax04}. SphereSort(${\bf X}_{q_r}$) returns data vectors ${\bf x}_i \in {\bf X}_{q_r}$ sorted in descending order of distance in the kernel space from the center of the SVDD hypersphere. CheckPoint(${\bf x}_i, \Psi$) returns TRUE if ${\bf x}_i$ is an approximate extreme point of the set $\Psi$ in kernel space. The operator $A \backslash B$ indicates a set operation that returns the set of the members of $A$ excluding $A \cap B$. The matrix ${\bf X}_{q_r}^*$ contains the approximate extreme points of ${\bf X}_{q_r}$ and $\overline{\beta_{q_r}}$ is a $|{\bf X}_{q_r}^*|$ sized vector.
\floatname{algorithm}{ }
\begin{algorithm}[h]
 \renewcommand{\thealgorithm}{}
\caption{[${\bf X}_{q_r}^*,\overline{\beta_{q_r}}$] = DeriveAE(${\bf X}_{q_r}$)}
\begin{enumerate} \addtolength{\itemsep}{-.35\baselineskip}
\item Initialize: ${\bf X}_{q_r}^*$ = SphereSet(${\bf X}_{q_r}$) and $\Psi = \emptyset$
\item Set $\zeta$ = SphereSort(${\bf X}_{q_r}\backslash {\bf X}_{q_r}^*$)
\item For each ${\bf x}_i$ taken in order from $\zeta$, call the routine CheckPoint(${\bf x}_i,{\bf X}_{q_r}^* \cup \Psi$)
\subitem If it returns $FALSE$, then set $\Psi = \Psi \cup {\bf x}_i$
\item For each ${\bf x}_i \in \Psi$, execute CheckPoint(${\bf x}_i,{\bf X}_{q_r}^* \cup \{\Psi \backslash {\bf x}_i\}$)
\subitem If it returns $FALSE$, set ${\bf X}_{q_r}^* = {\bf X}_{q_r}^* \cup {\bf x}_i$
\item Initialize a matrix $\Gamma$ of size $|{\bf X}_{q_r}|$x$|{\bf X}_{q_r}^*|$ with all elements set to 0
\subitem Set $\mu^k _k = 1 \;\forall {\bf x}_k \in {\bf X}_{q_r}^*$, where $\mu^i _j$ is the element in the $i^{th}$ row and $j^{th}$ column of $\Gamma$
\item For each ${\bf x}_i \in {\bf X}_{q_r}$ and ${\bf x}_i \not\in {\bf X}_{q_r}^*$, execute CheckPoint(${\bf x}_i,{\bf X}_{q_r}^*$)
\subitem Set the $i^{th}$ row of $\Gamma$ = $\overline{\mu ^i}$, where $\overline{\mu ^i}$ is the result of CheckPoint(${\bf x}_i,{\bf X}_{q_r}^*$)
\item For j = 1 {\bf{\emph to}} $|{\bf X}_{q_r}^*|$
\subitem Set $\beta_{q_r} ^ j = \overset{|{\bf X}_{q_r}|}{\underset{k = 1}{\sum}} \mu^k _j$
\end{enumerate}
\end{algorithm}

CheckPoint(${\bf x}_i, \Psi$) is computed by solving the following quadratic optimization problem:
\begin{align*}
&\underset {\overline{\mu ^i}}{\textbf{min }} p({\bf x}_i,\Psi) = \|\phi({\bf x}_i) - \overset{|\Psi|}{\underset{t = 1}{\sum}}\mu_t ^i \phi({\bf x}_t)\|^2\\
&\text{ s.t. } {\bf x}_t \in \Psi, 0 \le \mu_t ^i \le 1 \text{ and }  \overset{|\Psi|}{\underset{t = 1}{\sum}} \mu_t ^i = 1
\end{align*}
where $\|\phi({\bf x}_i) - \overset{|\Psi|}{\underset{t = 1}{\sum}}\mu_t ^i \phi({\bf x}_t)\|^2 = K({\bf x}_t,{\bf x}_t) + \overset{|\Psi|}{\underset{t = 1}{\sum}}\overset{|\Psi|}{\underset{s = 1}{\sum}}\mu_t ^i\mu_s ^i K({\bf x}_t, {\bf x}_s) - 2 \overset{|\Psi|}{\underset{t = 1}{\sum}}\mu_t ^i K({\bf x}_i, {\bf x}_t)$. If the optimized value of $p({\bf x}_i,\Psi)\le \epsilon$, CheckPoint(${\bf x}_i, \Psi$) returns TRUE and otherwise it returns FALSE. It can be seen that the formulation of $p({\bf x}_i,\Psi)$ is similar to (\ref{eq:AEX}). The value of $\overline{\mu ^i}$ computed by CheckPoint(${\bf z}_i, \Psi_0$), is used in step 6 of DeriveAE. 

Now we compute the time complexity of DeriveAE. We use the fact that the optimization problem in CheckPoint(${\bf x}_i, \Psi$) is essentially the same as the dual optimization problem of SVM given in (\ref{eq:SVC_dual}). Since DeriveAE solves several SVM training problems in steps 1,3,4, and 6, it is necessary to know the training time complexity of a SVM. As any SVM solver method can be used, we denote the training time complexity of each step of DeriveAE that solves an SVM problem as $O(S(A_{q_r}))$ \footnote{For SMO based implementations, such as the implementation we used for Section \ref{sec:exp}, $S(A) = O(A^2)$}. Here $A_{q_r}$ is the largest value of ${\bf X}_{q_r}^* \cup \Psi$ during the run of DeriveAE(${\bf X}_{q_r}$). This enables us to derive a generic expression for the complexity of DeriveAE, independent of the SVM solver method used. Hence the time complexity of step 1 is $O(S(A_{q_r}))$. The time complexity of steps 3, 4 and 6 are $O(V\;S(A_{q_r}))$, $O(A_{q_r}\;S(A_{q_r}))$, and $O(A_{q_r}\;S(A_{q_r}))$ respectively. The time complexity of step 2 is $O(V \;|\Psi_1| + V \text{ log}_2 V)$, where $\Psi_1$ = SphereSet(${\bf X}_{q_r}$). Hence the time complexity of DeriveAE is $O(V \;|\Psi| + V \text{ log}_2 V + V\;S(A_{q_r}) + A_{q_r}\;S(A_{q_r}))$. Since $|\Psi_1|$ is typically very small and $A_{q_r} \le V$, we denote the time complexity of DeriveAE by $O(V \text{ log}_2 V + V\;S(A_{q_r}))$.

\subsection{Combining all the methods to compute ${X}^*$}
To derive ${X}^*$, it is required to first rearrange ${\bf X}$, so that data vectors from each class are grouped together as ${\bf X} = \{{\bf X}^+,{\bf X}^-\}$. Here ${\bf X}^+  = \{{\bf x}_i: y_i = 1, {\bf x}_i \in {\bf X}\} \text{ and } {\bf X}^-  = \{{\bf x}_i: y_i = -1, {\bf x}_i \in {\bf X}\}$. Then the selected segregation methods are run on ${\bf X}^+$ and ${\bf X}^-$ separately. The algorithm DeriveRS given below, combines all the algorithms defined earlier in this section with a few additional steps, to compute the representative set of ${\bf X}$. The complexity of DeriveRS \footnote{We present DeriveRS as one algorithm in spite of its two variants that use FLS1 or FLS2, for simplicity and to conserve space.} can easily be computed by summing the complexities of its steps. The complexity of steps 1 and 6 is O(N). The complexity of step 2 is $O(N)$ if FLS1 is run or $O(N \;\text{log}_2 \frac{N}{P})$ if FLS2 is run. In step 3, the $O(\frac{NP}{V})$ method SLS is run. In steps 4 and 5, DeriveAE is run on all the subsets ${\bf X}_{q_r}$ giving a total complexity of $O(N \text{ log}_2 V + V\overset{Q}{\underset{q = 1}{\sum}} \overset{R}{\underset{r = 1}{\sum}} S(A_{q_r}))$. Here we use the fact that the number of subsets ${\bf X}_{q_r}$ is $O(\frac{N}{V})$. Thus the complexity of DeriveRS is $O(N(\frac{P}{V} + \text{ log}_2 V) + V\overset{Q}{\underset{q = 1}{\sum}} \overset{R}{\underset{r = 1}{\sum}} S(A_{q_r}))$ when FLS1 is used and $O(N (\text{log}_2 \frac{N}{P} + \frac{P}{V} + \text{log}_2 V) + V\overset{Q}{\underset{q = 1}{\sum}} \overset{R}{\underset{r = 1}{\sum}} S(A_{q_r}))$ when FLS2 is used.

\floatname{algorithm}{ }
\begin{algorithm}[h]
 \renewcommand{\thealgorithm}{}
\caption{[${\bf X}^*,{\bf Y}^*,\overline{\beta}$] = DeriveRS(${\bf X}$,${\bf Y}$,P,V)}
\begin{enumerate}
  \item Set ${\bf X}^+ = \{{\bf x}_i: {\bf x}_i \in {\bf X}, y_i = 1\}$ and ${\bf X}^- = \{{\bf x}_i: {\bf x}_i \in {\bf X}, y_i = -1\}$
  \item Run [${\bf X}^+, \Delta^+$] = FLS(${\bf X}^+$,P) and [${\bf X}^-, \Delta^-$] = FLS(${\bf X}^-$,P), where FLS is FLS1 or FLS2
	\item Run [${\bf X}^+, \Delta^+ _2$] = SLS(${\bf X}^+$,V,$\Delta^+$) and [${\bf X}^-, \Delta^- _2$] = SLS(${\bf X}^-$,V,$\Delta^-$)
  \item Using $\Delta^+ _2$, identify each subset ${\bf X}_{q_r}$ of ${\bf X}^+$ and run [${\bf X}_{q_r}^*,\overline{\beta_{q_r}}$] = DeriveAE(${\bf X}_{q_r}$)
  \subitem Set $N^{+*} =$ sum of number of data vectors in all ${\bf X}_{q_r}^*$ derived from ${\bf X}^+$
  \item Using $\Delta^- _2$, identify each subset ${\bf X}_{q_r}$ of ${\bf X}^-$ and run [${\bf X}_{q_r}^*,\overline{\beta_{q_r}}$] = DeriveAE(${\bf X}_{q_r}$)
  \subitem Set $N^{-*} =$ sum of number of data vectors in all ${\bf X}_{q_r}^*$ derived from ${\bf X}^-$  
  \item Combine in the same order, all ${\bf X}_{q_r}^*$ to obtain ${\bf X}^*$ and all $\overline{\beta_{q_r}}$ to obtain $\overline{\beta}$
  \subitem Set ${\bf Y}^* = \{y_i: y_i = 1 \text{ for } i = 1,2,..,N^{+*} \text{; and }y_i = -1 \text{ for } i = 1+N^{+*}, 2+N^{+*},..,N^{-*}+N^{+*}\}$
\end{enumerate}
\end{algorithm}

\section{Experiments}\label{sec:exp}
We focused our experiments on an SMO \citep{Fan05} based implementation of AESVM and DeriveRS. We evaluated the classification performance of AESVM using the nine datasets, described below. Next, we present an evaluation of the algorithm DeriveRS, followed by an evaluation of AESVM.%We trained the methods on the Gaussian kernel $K({\bf x}_i,{\bf x}_j) = e^{(-g \|{\bf x}_i -{\bf x}_j\|^2)}$, where $g$ is a kernel hyper-parameter.

\subsection{Datasets}
Nine datasets of varied size, dimensionality and density were used to evaluate DeriveRS and our AESVM implementation. For datasets D2, D3 and D4, we performed five fold cross validation. We did not perform five fold cross-validation on the other datasets, because they have been widely used in their native form with a separate training and testing set. 

\begin{description}
\item[D1: ]{\it KDD'99 intrusion detection dataset\footnote{\url{http://archive.ics.uci.edu/ml/datasets/KDD+Cup+1999+Data}}}- This dataset is available as a training set of 4898431 data vectors and a testing set of 311027 data vectors, with forty one features ($D = 41$). As described in \citet{Tavallaee09}, a huge portion of this dataset is comprised of repeated data vectors. Experiments were conducted only on the distinct data vectors. The number of distinct training set vectors was $N = 1074974$ and the number of distinct testing set vectors was $N = 77216$. The training set density = 33\%.

\item[D2: ]{\it Localization data for person activity\footnote{\url{http://archive.ics.uci.edu/ml/datasets/Localization+Data+for+Person+Activity}} }- This dataset has been used in a study on agent-based care for independent living \citep{Kaluza10}. It has $N = 164860$ data vectors of seven features. It is comprised of continuous recordings from sensors attached to five people and can be used to predict the activity that was performed by each person at the time of data collection. In our experiments we used this dataset to validate a binary problem of classifying the activities 'lying' and 'lying down' from the other activities. Features 3 and 4, that gives the time information, were not used in our experiments. Hence for this dataset $D = 5$. The dataset density = 96\%.

\item[D3: ]{\it  Seizure detection dataset}- This dataset has $N = 982863$ data vectors, three features ($D = 3$) and density = 100\%. It is comprised of continuous EEG recordings from rats induced with status epilepticus and is used to evaluate algorithms that classify seizure events from seizure-free EEG. An important characteristic of this dataset is that it is highly unbalanced, the total number of data vectors corresponding to seizures is minuscule compared to the remaining data. Details of the dataset can be found in \citet{Nandan10}, where it is used as dataset A. 

\item[D4: ]{\it Forest cover type dataset\footnote{\url{http://archive.ics.uci.edu/ml/datasets/Covertype}}}- This dataset has $N = 581012$ data vectors and fifty four features ($D = 54$) and density = 22\%. It is used to classify the forest cover of areas of 30mx30m size into one of seven types. We followed the method used in \citet{Collobert02}, where a classification of forest cover type 2 from the other cover types was performed. 

\item[D5 :] {\it IJCNN1 dataset} \footnote{\url{http://www.csie.ntu.edu.tw/~cjlin/libsvmtools/datasets/binary.html}}- This dataset was used in IJCNN 2001 generalization ability challenge \citep{Chang01}. The training set and testing set have 49990 ($N = 49990$) and 91701 data vectors respectively. It has 22 features ($D = 22$) and training set density = 59\%

\item[D6 :] {\it Adult income dataset} \footnote{\url{http://www.csie.ntu.edu.tw/~cjlin/libsvmtools/datasets/binary.html}}- This dataset derived from the 1994 Census database, was used to classify incomes over \$50000 from those below it. The training set has $N = 32561$ with $D = 123$ and density = 11\%, while the testing set has 16281 data vectors. The data is pre-processed as described in \citet{Platt99}. 

\item[D7 :] {\it Epsilon dataset} \footnote{\url{http://www.csie.ntu.edu.tw/~cjlin/libsvmtools/datasets/binary.html}}- This is a dataset that was used for 2008 Pascal large scale learning challenge and in \citet{Yuan11}. It is comprised of 400000 data vectors that are 100\% dense with $D = 2000$. Since this is too large for our experiments, we used the first 10\% of the training set giving $N = 40000$. The testing set has 100000 data vectors.

\item[D8 :] {\it MNIST character recognition dataset} \footnote{\url{http://www.csie.ntu.edu.tw/~cjlin/libsvmtools/datasets/multiclass.html}}- The widely used dataset \citep{Lecun98} of hand written characters has a training set of $N = 60000$, $D = 780$ and density = 19\%. We performed the binary classification task of classifying the character '0' from the others. The testing set has 10000 data vectors.

\item[D9 :] {\it w8a dataset} \footnote{\url{http://www.csie.ntu.edu.tw/~cjlin/libsvmtools/datasets/binary.html}}- This artificial dataset used in \citet{Platt99} was randomly generated and has $D = 300$ features. The training set has $N = 49749$ with a density = 4\% and the testing set has 14951 data vectors.

\end{description}

\subsection{Evaluation of DeriveRS} \label{sec:derRS}
\begin{figure}[h!]
\centering
\includegraphics[scale = 0.38]{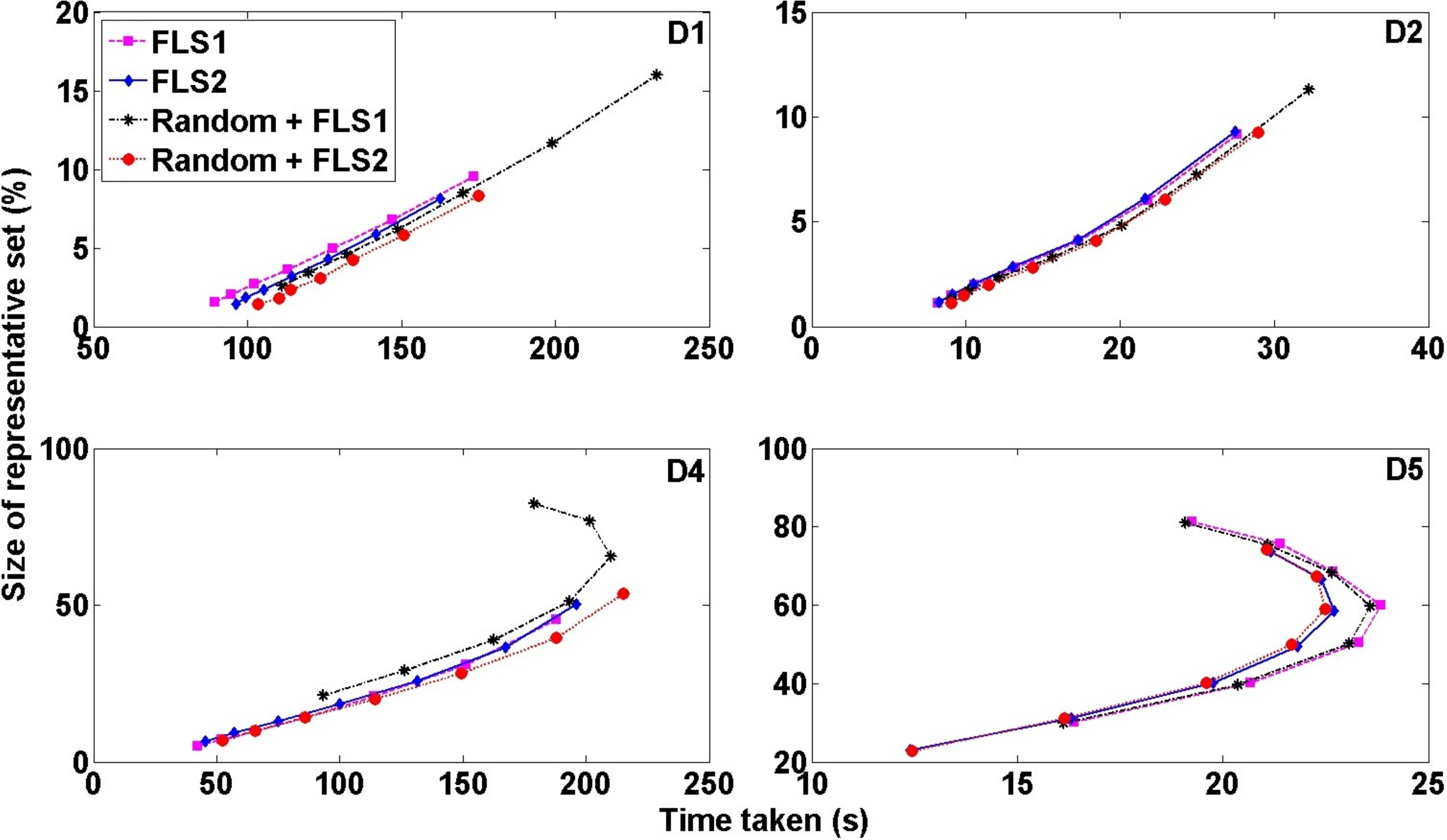}
\caption{Performance of variants of DeriveRS with $g = [2^{-4},2^{-3},2^{-2},...,2^2]$, for datasets D1, D2, D4, and D5. The results of DeriveRS with FLS1 and FLS2, after randomly reordering the datasets are shown as Random+FLS1 and Random+FLS2, respectively}\label{fig:resFLS}
\end{figure}

We began our experiments with an evaluation of the algorithm DeriveRS, described in Section \ref{sec:derRP}. The performances of the two methods FLS1 and FLS2 were compared first. We ran DeriveRS on D1, D2, D4 and D5 with the parameters $P = 10^4, V = 10^3, \epsilon = 10^{-3}$, and $g = [2^{-4},2^{-3},2^{-2},...,2^2]$, first with FLS1 and then FLS2. For D2, DeriveRS was run on the entire dataset for this particular experiment, instead of performing five fold cross-validation. This was done because, D2 is a small dataset and the difference between the two first level segregation methods can be better observed when the dataset is as large as possible. The relatively small value of $P = 10^4$ was also chosen considering the small size of D2 and D5. To evaluate the effectiveness of FLS1 and FLS2, we also ran DeriveRS with FLS1 and FLS2 after randomly reordering each dataset. The results are shown in Figure \ref{fig:resFLS}. 

For datasets D1 and D5, FLS2 gave smaller representative sets in a shorter duration than FLS1. As expected, for the relatively homogeneous dataset D2, FLS1 and FLS2 gave similar results, with FLS2 giving slightly larger representative sets. Dataset D4 was seen to have much smaller representative sets with FLS1 than with FLS2. The results of DeriveRS obtained after randomly rearranging the datasets, indicate the utility of FLS2. For all the datasets, the results of FLS2 after random reordering was seen to be significantly better than the results of FLS1 after random rearrangement. Hence we can infer that the good results obtained with FLS2 are not caused by any pre-existing order in the datasets. After D2 and D4 were randomly rearranged a sharp increase was observed in representative set sizes and computation times for DeriveRS with FLS1. This indicates the importance of dataset homogeneity to the performance of FLS1. The results indicated for randomized experiments on DeriveRS are the averages of five repetitions.

Next we investigated the impact of changes in the values of the parameters $P$ and $V$ on the performance of DeriveRS. All combinations of $P = \{10^4, 5\text{x}10^4,10^5,2\text{x}10^5\}$ and $V = \{10^2,5\text{x}10^2,10^3,2\text{x}10^3,3\text{x}10^3\}$ were used to compute the representative set of D1. The computations were performed for $\epsilon = 10^{-3}$ and $g = 1$. The method FLS2 was used for the first level segregation in DeriveRS. The results are shown in Table \ref{tb:RP2}. As expected for an algorithm of time complexity $O(N (\text{log}_2 \frac{N}{P} + \frac{P}{V} + \text{log}_2 V) + V\overset{Q}{\underset{q = 1}{\sum}} \overset{R}{\underset{r = 1}{\sum}} S(A_{q_r}))$,  the computation time was generally observed to increase for an increase in the value of $V$ or $P$. It should be noted that our implementation of DeriveRS was based on SMO and hence $S(A_{q_r}) = O(A_{q_r} ^2)$. In some cases the computation time decreased when $P$ or $V$ increased. This is caused by a decrease in the value of $O(\overset{Q}{\underset{q = 1}{\sum}} \overset{R}{\underset{r = 1}{\sum}} A_{q_r} ^2)$, which is inferred from the observed decrease of the size of the representative set $M$ ($M \approx \overset{Q}{\underset{q = 1}{\sum}} \overset{R}{\underset{r = 1}{\sum}} A_{q_r}$). A sharp decrease in $M$ was observed when $V$ was increased. The impact of increasing $P$ on the size of the representative set was found to be less drastic. This observation indicates that DeriveAE selects fewer approximate extreme points when $V$ is larger. 

\begin{table}[h!]
\begin{center}
\begin{tabular}{|c!{\vrule width 1pt}c|c|c|c|c|}\hline
\multicolumn{6}{|c|}{$\frac{M}{N}$x100\%  (Computation time in seconds)}\\ \noalign{\hrule height 1pt}
$P$&$V = 10^2$&$V = 5\text{x}10^2$&$V = 10^3$&$V = 2\text{x}10^3$&$V = 3\text{x}10^3$\\ \hline\hline
$10^4$& 10.7(27) & 6.1(67) & 5.1(131) & 4.5(258) & 4.3(338) \\ \hline
$5\text{x}10^4$& 9.9(78) & 5.3(72) & 4.4(130) & 3.9(249) & 3.7(351) \\ \hline
$10^5$& 9.8(142) & 5.2(83) & 4.3(134) & 3.7(242) & 3.5(352) \\ \hline
$2\text{x}10^5$& 9.8(254) & 5.1(104) & 4.2(144) & 3.7(240) & 3.4(355) \\ \hline
\end{tabular}
\end{center}
\caption{The impact of varying $P$ and $V$ on the result of DeriveRS}
\label{tb:RP2}
\end{table}

As described in Section \ref{sec:compAE}, we compared several SVM training algorithms with our implementation of AESVM. We performed a grid search with all combinations of the SVM hyper-parameters $C' = \{2^{-4},2^{-3},...,2^6,2^7\}$ and $g = \{ 2^{-4},2^{-3},2^{-2},...,2^1,2^2\}$. The hyper-parameter $C'$ is related to the hyper-parameter $C$ as $C' = \frac{C}{N}$. We represent the grid in terms of $C'$ as it is used in several SVM solvers such as LIBSVM, LASVM, CVM and BVM. Furthermore, the use of $C'$ enables the application of the same hyper-parameter grid to all datasets. To train AESVM with all the hyper-parameter combinations in the grid, the representative set has to be computed using DeriveRS for all values of kernel hyper-parameter $g$ in the grid. This is because the kernel space varies when the value of $g$ is varied. For all the computations, the input parameters were set as $P= 10^5$ and $V= 10^3$. The first level segregation in DeriveRS was performed using FLS2. Three values of the tolerance parameter $\epsilon$ were investigated, $\epsilon = 10^{-3},10^{-4} \text{ or }10^{-5}$. 

The results of the computation for datasets D1 - D5, are shown in the Table \ref{tb:RP1}. The percentage of data vectors in the representative set was found to increase with increasing values of $g$. This is intuitive, as when $g$ increases the distance between the data vectors in kernel space increases. With increased distances, more data vectors ${\bf x}_i$ become approximate extreme points. The increase in the number of approximate extreme points with $g$ causes the rising trend of computation time shown in Table \ref{tb:RP1}. For a decrease in the value of $\epsilon$, $M$ increases. This is because, for smaller $\epsilon$ fewer ${\bf x}_i$ would satisfy the condition: optimized $p({\bf x}_i,\Psi)\le \epsilon$ in CheckPoint(${\bf x}_i, \Psi$). This results in the selection of a larger number of approximate extreme points in DeriveAE.

\begin{table}[h!]
\begin{center}
\begin{tabular}{|@{}p{0.6cm}!{\vrule width 1pt}@{}p{1.2cm}!{\vrule width 1pt}@{}p{1.4cm}|@{}p{1.4cm}|@{}p{1.6cm}|@{}p{1.4cm}|@{}p{1.6cm}|@{}p{1.6cm}|@{}p{1.6cm}|}\hline
\multicolumn{9}{|c|}{$\frac{M}{N}$x100\%  (Computation time in seconds)}\\ \noalign{\hrule height 1pt}
$\epsilon$&Dataset&g = $\frac{1}{2^4}$&g = $\frac{1}{2^3}$&g = $\frac{1}{2^2}$&g = $\frac{1}{2}$&g = $1$&g = $2^{1}$&g = $2^{2}$\\ \hline\hline
\multirow{5}{*}{$10^{-3}$}& D1 & 1.5(98) & 1.9(104) & 2.4(110) & 3.2(119) & 4.3(132) & 5.9(148) & 8.1(168)  \\ \cline{2-9}
& D2 & 1.2(7) & 1.5(8) & 2(9) & 2.8(11) & 4.1(15) & 6(18) & 9.2(23) \\ \cline{2-9}
& D3 & 0.6(37) & 0.6(37) & 0.6(36) & 0.6(36) & 0.5(37) & 0.6(37) & 0.6(39) \\ \cline{2-9}
& D4 & 4.3(45) & 6.4(57) & 9.4(74) & 13.9(103) & 20.7(139) & 30.7(178) & 44.8(216) \\ \cline{2-9}
& D5 & 4.5(7) & 8.3(9) & 14(11) & 21.8(14) & 31.8(18) & 43.7(21) & 54.9(22) \\ \noalign{\hrule height 1pt}
\multirow{5}{*}{$10^{-4}$}& D1 & 3(136) & 4(159) & 5.3(191) & 7.2(240) & 9.9(297) & 13.3(362) & 17.4(435)  \\ \cline{2-9}
& D2 & 2.8(12) & 3.8(18) & 5(27) & 6.8(37) & 9.3(44) & 13.5(44) & 19.9(82) \\ \cline{2-9}
& D3 & 0.5(36) & 0.6(37) & 0.6(38) & 0.7(39) & 0.8(41) & 0.9(43) & 1.1(47) \\ \cline{2-9}
& D4 & 13.5(135) & 18.3(211) & 24.9(300) & 34.2(400) & 47.7(493) & 63.5(513) & 74.4(445) \\ \cline{2-9}
& D5 & 20.1(16) & 27.9(22) & 37.4(27) & 47.6(31) & 57.3(34) & 66(34) & 74(34) \\ \noalign{\hrule height 1pt}
\multirow{5}{*}{$10^{-5}$}& D1 & 7(316) & 9.3(425) & 12.2(552) & 15.7(726) & 19.6(926) & 24.2(1112) & 28.9(1235)  \\ \cline{2-9}
& D2 & 6.2(59) & 7.8(87) & 9.8(98) & 13(109) & 18.3(138) & 25.6(187) & 34.3(235) \\ \cline{2-9}
& D3 & 0.7(39) & 0.8(42) & 0.9(45) & 1.1(50) & 1.4(59) & 1.7(73) & 2.2(100) \\ \cline{2-9}
& D4 & 30.7(607) & 39.5(814) & 51.9(1051) & 66(1171) & 75.1(1044) & 77.8(839) & 78.4(649) \\ \cline{2-9}
& D5 & 43.3(50) & 51.8(58) & 60.3(62) & 67.7(63) & 73.8(59) & 78.7(52) & 81.8(44) \\ \hline
\end{tabular}
\end{center}
\caption{The percentage of the data vectors in ${\bf X}^*$ (given by $\frac{M}{N}$x100) and its computation time for datasets D1-D5}
\label{tb:RP1}
\end{table}

The results of applying DeriveRS to the high-dimensional datasets D6-D9 are shown in Table \ref{tb:RP4}. It was observed that $\frac{M}{N}$ was much larger for D6-D9 than for the other datasets. We computed the representative set with $\epsilon= 10^{-3}$ only, as for smaller values of $\epsilon$ we expect the representative set to be close to 100\% of the training set. The increasing trend of the size of the representative set with increasing $g$ values can be observed in Table \ref{tb:RP4} also.

\begin{table}[h!]
\begin{center}
\begin{tabular}{|@{}p{1.2cm}!{\vrule width 1pt}@{}p{1.6cm}|@{}p{1.6cm}|@{}p{1.6cm}|@{}p{1.6cm}|@{}p{1.6cm}|@{}p{1.6cm}|@{}p{1.4cm}|}\hline
\multicolumn{8}{|c|}{$\frac{M}{N}$x100\%  (Computation time in seconds)}\\ \noalign{\hrule height 1pt}
Dataset&g = $\frac{1}{2^4}$&g = $\frac{1}{2^3}$&g = $\frac{1}{2^2}$&g = $\frac{1}{2}$&g = $1$&g = $2^{1}$&g = $2^{2}$\\ \hline\hline
D6 & 69.3(19) & 70.4(19) & 73.4(19) & 80.3(14) & 83.9(9) & 84(8) & 87.9(8) \\ \hline
D7 & 84.4(1077) & 84.6(1089) & 84.9(1069) & 85.6(1085) & 86.9(1079) & 89.9(1032) & 94.7(818) \\ \hline
D8 & 90(131) & 96.6(94) & 98.8(78) & 99.5(72) & 100(70) & 100(71) & 100(63)  \\ \hline
D9 & 60.8(34) & 62.9(36) & 67(30) & 70.8(21) & 72.7(16) & 75.2(14) & 76.7(15) \\ \noalign{\hrule height 1pt}
\end{tabular}
\end{center}
\caption{The percentage of data vectors in ${\bf X}^*$ and its computation time for datasets D6-D9 with $\epsilon= 10^{-3}$}
\label{tb:RP4}
\end{table}

\subsection{Comparison of AESVM to SVM solvers} \label{sec:compAE}

To judge the accuracy and efficiency of AESVM, its classification performance was compared with the SMO implementation in LIBSVM, ver. 3.1. We chose LIBSVM because it is a state-of-the-art SMO implementation that is routinely used in similar comparison studies. To compare the efficiency of AESVM to other popular approximate SVM solvers we chose CVM, BVM, LASVM, $\text{SVM}^{\text{perf}}$, and RfeatSVM. A description of these methods is given in Section \ref{sec:relWork}. We chose these methods because they are widely cited, their software implementations are freely available and other studies \citep{Shwartz11} have reported fast SVM training using some of these methods. LASVM is also an efficient method for online SVM training. However, since we do not investigate online SVM learning in this paper, we did not test the online SVM training performance of LASVM. We compared AESVM with CVM and BVM even though they are L2-SVM solvers, as they has been reported to be faster alternatives to SVM implementations such as LIBSVM.

The implementation of AESVM and DeriveRS were built upon the LIBSVM implementation. All methods except $\text{SVM}^{\text{perf}}$ were allocated a cache of size 600 MB. The parameters for DeriveRS were $P = 10^5$ and $V= 10^3$, and the first level segregation was performed using FLS2. To reflect a typical SVM training scenario, we performed a grid search with all eighty four combinations of the SVM hyper-parameters $C' = \{2^{-4},2^{-3},...,2^6,2^7\}$ and $g = \{2^{-4},2^{-3},2^{-2},...,2^1,2^2\}$. As mentioned earlier, for datasets D2, D3 and D4, five fold cross-validation was performed. The results of the comparison have been split into sub-sections given below, due to the large number of SVM solvers and datasets used.%The authors have made all the relevant code available for download [give cite of URL]. 

\subsubsection{Comparison to CVM, BVM, LASVM and LIBSVM}

\begin{figure}[h!]
\centering
\includegraphics[scale = 0.52]{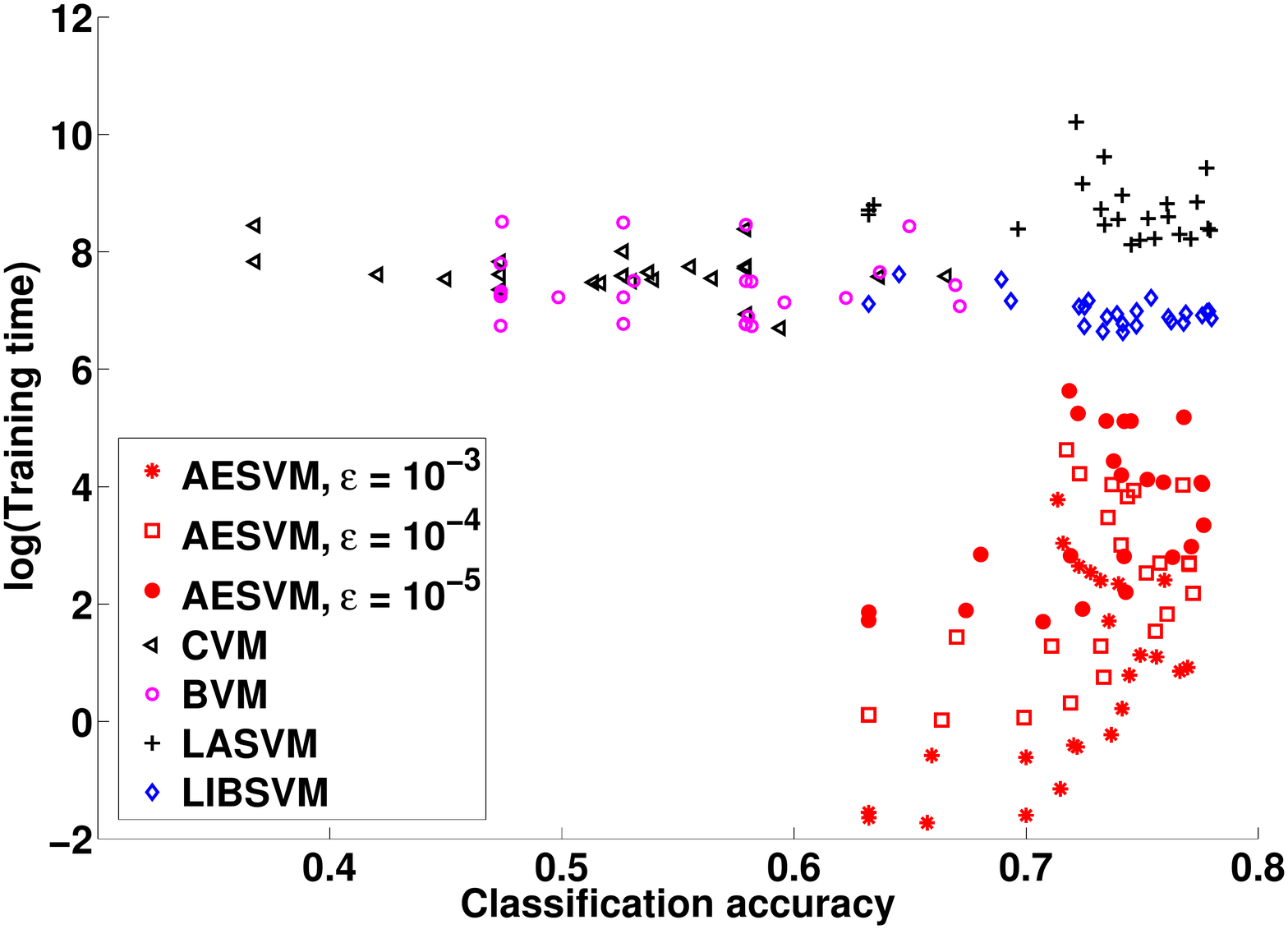}
\caption{Plot of training time against classification accuracy of the SVM algorithms on D2}\label{fig:paRes1}
\end{figure}

\begin{figure}[h!]
\centering
\includegraphics[scale = 0.52]{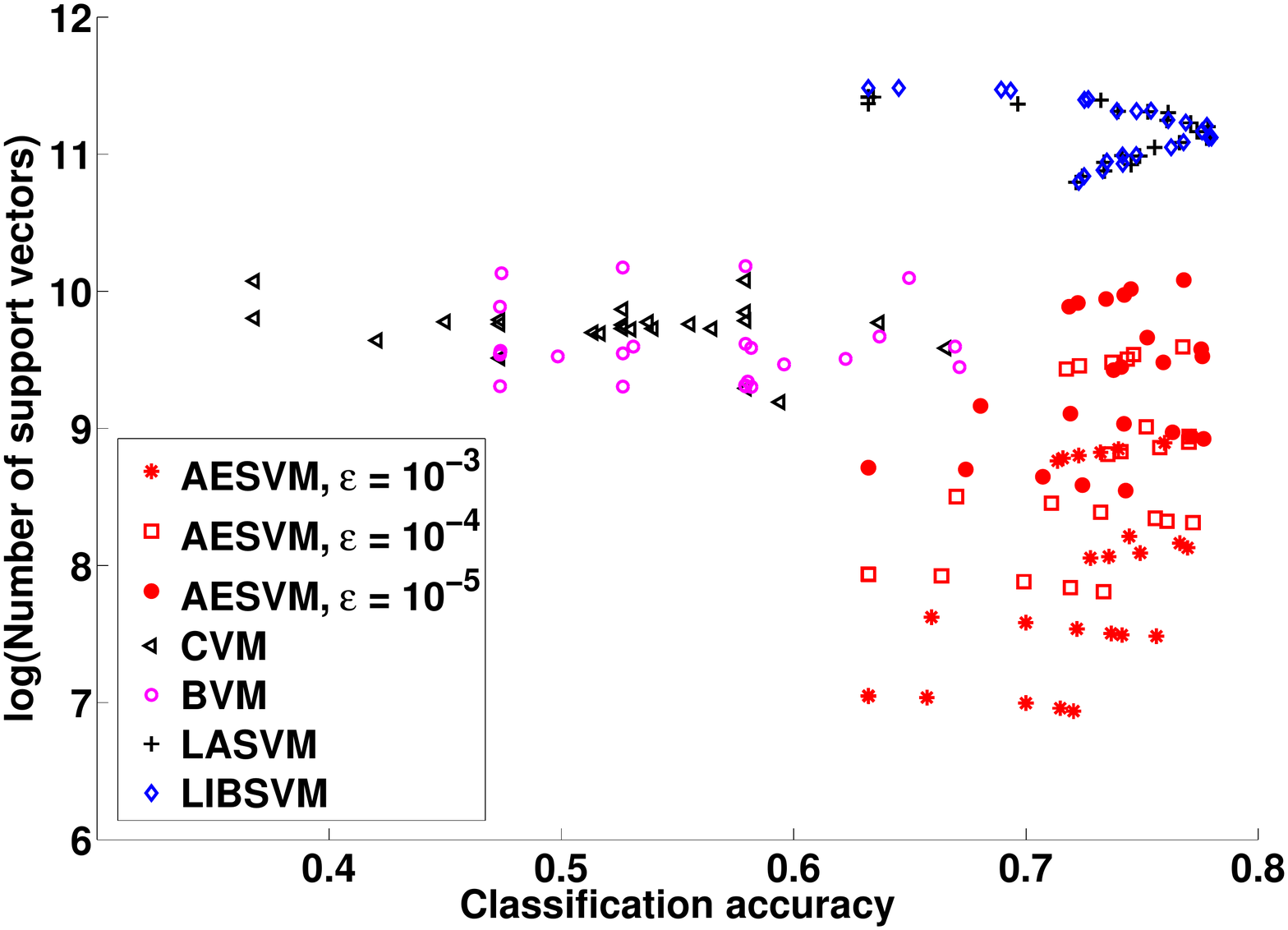}
\caption{Plot of classification time, represented by the number of support vectors, against classification accuracy of the SVM algorithms on D2}\label{fig:paRes2}
\end{figure}

First we present the results of the performance comparison for D2 in Figures \ref{fig:paRes1} and \ref{fig:paRes2}. For ease of representation, only the results of grid points corresponding to combinations of $C' = \{2^{-4},2^{-2},1,2^{2},2^{4},2^6\}$ and $g = \{2^{-4},2^{-2},1,2^2\}$ are shown in Figures \ref{fig:paRes1} and \ref{fig:paRes2}. Figure \ref{fig:paRes1} shows the graph between training time and classification accuracy for the five algorithms. Figure \ref{fig:paRes2} shows the graph between the number of support vectors and classification accuracy. We present classification accuracy as the ratio of the number of correct classifications to the total number of classifications performed. Since the classification time of an SVM algorithm is directly proportional to the number of support vectors, we represent it in terms of the number of support vectors. It can be seen that, AESVM generally gave more accurate results for a fraction of the training time of the other algorithms, and also resulted in less classification time. The training time and classification times of AESVM increased when $\epsilon$ was reduced. This is expected given the inverse relation of $M$ to $\epsilon$ shown in Tables \ref{tb:RP1} and \ref{tb:RP4}. The variation in accuracy with $\epsilon$ is not very noticeable. 

Figures \ref{fig:paRes1} and \ref{fig:paRes2} indicate that AESVM gave better results than the other algorithms for SVM training and classification on D2, in terms of standard metrics. To present a more quantitative and easily interpretable comparison of the algorithms, we define the five performance metrics given below. These metrics combine the results of all runs of each algorithm into a single value, for each dataset. For these metrics we take LIBSVM as a baseline of comparison, as it gives the most accurate solution among the tested methods. Furthermore, an important objective of these experiments is to show the similarity of the results of AESVM and LIBSVM. In the description given below, $\mathbb{F}$ can refer to any or any approximate SVM algorithm such as AESVM, CVM, LASVM etc.

\begin{enumerate}\addtolength{\itemsep}{-.35\baselineskip}
	\item {\it Root mean squared error of classification accuracy, $RMSE$}: The similarity of the solution of $\mathbb{F}$ to LIBSVM, in terms of its classification accuracy, is indicated by:
	\begin{equation*}
	RMSE = \left(\frac{1}{RS}{\overset{R}{\underset{r = 1}{\sum}} \overset{S}{\underset{s = 1}{\sum}} (CL^r _s - C\mathbb{F}^r _s)^2}\right)^{0.5}
	\end{equation*}
		Here $CL^r _s$ and $C\mathbb{F}^r _s$ are the classification accuracy of LIBSVM and $\mathbb{F}$ respectively, in the $s^{th}$ cross-validation fold with the $r^{th}$ set of hyper-parameters of grid search.
		
	\item{\it Expected training time speedup, $ETS$}: The expected speedup in training time is indicated by:
	\begin{equation*}
	ETS = \frac{1}{RS}{\overset{R}{\underset{r = 1}{\sum}} \overset{S}{\underset{s = 1}{\sum}} \frac{TL^r _s}{T\mathbb{F}^r _s}}
	\end{equation*}
		Here $TL^r _s$ and $T\mathbb{F}^r _s$ are the training times of LIBSVM and $\mathbb{F}$ respectively.
		
	\item{\it Overall training time speedup, $OTS$}: It indicates overall training time speedup for the entire grid search with cross-validation, including the time taken to compute the representative set. The total time taken by DeriveRS to compute the representative set  for all values of $g$ is represented as $\mathbb{TX}^*$. For methods other than AESVM, $\mathbb{TX}^* = 0$.
	\begin{equation*}
	OTS = \frac{\overset{R}{\underset{r = 1}{\sum}} \overset{S}{\underset{s = 1}{\sum}}{TL^r _s}}{ \overset{R}{\underset{r = 1}{\sum}} \overset{S}{\underset{s = 1}{\sum}} T\mathbb{F}^r _s + \mathbb{TX}^*}
	\end{equation*}

	\item{\it Expected classification time speedup, $ECS$}: The expected speedup in classification time is indicated by:
	\begin{equation*}
	ECS = \frac{1}{RS}{\overset{R}{\underset{r = 1}{\sum}} \overset{S}{\underset{s = 1}{\sum}} \frac{NL^r _s}{N\mathbb{F}^r _s}}
	\end{equation*}
		Here $NL^r _s$ and $N\mathbb{F}^r _s$ are the number of support vectors in the solution of LIBSVM and $\mathbb{F}$ respectively.
	\item{\it Overall classification time speedup, $OCS$}: The overall speedup in classification time is indicated by:
	\begin{equation*}
	OCS = \frac{\overset{R}{\underset{r = 1}{\sum}} \overset{S}{\underset{s = 1}{\sum}}{NL^r _s}}{ \overset{R}{\underset{r = 1}{\sum}} \overset{S}{\underset{s = 1}{\sum}} N\mathbb{F}^r _s}
	\end{equation*}	
		
\end{enumerate}

\begin{table}[h!]
\begin{center}
\begin{tabular}{|c!{\vrule width 1pt}c!{\vrule width 1pt}p{1.5cm}|p{1.5cm}|p{1.5cm}|l|l|l|} \hline
Metric& Dataset& AESVM $\epsilon = 10^{-3}$ & AESVM $\epsilon = 10^{-4}$ & AESVM $\epsilon = 10^{-5}$ & CVM & BVM & LASVM\\ \noalign{\hrule height 1pt}
\multirow{5}{1.3cm}{$RMSE$ (x$10^2$)}&D1 & 0.28 & 0.16 & 0.21 & 0.44 & 0.6 & 0.12\\ \cline{2-8}
&D2 & 2.56 & 1.81 & 1.19 & 26.59 & 24.06 &2.18\\ \cline{2-8}
&D3 & 0.16 & 0.10 & 0.05 & 0.33 & 0.39 & 55.2  \\ \cline{2-8}
&D4 & 1.08 & 0.82 & 0.74 & 9.4 & 9.44 & \hspace{0.6cm}$-$ \\ \cline{2-8}
&D5 & 0.99 &0.39 & 0.23 & 0.74 & 0.84 & 0.13 \\ \noalign{\hrule height 1pt}
\multirow{5}{*}{$ETS$}&D1 & 451.5 & 145 & 41.7 & 8.9 & 28.6 & 0.8\\ \cline{2-8}
&D2 & 1614.7 & 289.6 & 62.8 & 0.7 & 0.8 & 0.2\\ \cline{2-8}
&D3 & 28012.3 & 14799.3 & 7573.8 & 60.4 & 76.8 & 0.9 \\ \cline{2-8}
&D4 & 103.1 & 13.8 & 3.4 & 8 & 6.6 & \hspace{0.6cm}$-$ \\ \cline{2-8}
&D5 & 40.2 & 5 & 2 & 0.3 & 0.5 & 0.6 \\ \noalign{\hrule height 1pt}
\multirow{5}{*}{$OTS$}&D1 & 92.1 & 34.2 & 9.5 & 6.2 & 21.6 & 0.8\\ \cline{2-8}
&D2 & 148.6 & 45.5 & 14.3 & 0.5 & 0.5 & 0.1\\ \cline{2-8}
&D3 & 968.5 & 800.6 & 514.4 & 23.9 & 22.8 & 0.5 \\ \cline{2-8}
&D4 & 11.9 & 4.1 & 2.2 & 6.2 & 4.4 & \hspace{0.6cm}$-$ \\ \cline{2-8}
&D5 & 5.2 & 2.5 & 1.5 & 0.2 & 0.3 & 0.5 \\ \noalign{\hrule height 1pt}
\multirow{5}{*}{$ECS$}&D1 & 4.8 & 3.6 & 2.8 & 1.2 & 2 & 1.1\\ \cline{2-8}
&D2 & 35.9 & 15.5 & 7.9 & 4.7 & 5 & 1\\ \cline{2-8}
&D3 & 48.7 & 25.8 & 13.4 & 0.4 & 0.6 & 0.6 \\ \cline{2-8}
&D4 & 8.4 & 3.3 & 1.8 & 12.4 & 12.1 & \hspace{0.6cm}$-$ \\ \cline{2-8}
&D5 & 4.3 & 1.9 & 1.4 & 0.8 & 1 & 1 \\ \noalign{\hrule height 1pt}
\multirow{5}{*}{$OCS$}&D1 & 3.8 & 3.1 & 2.5 & 1.1 & 1.9 & 1\\ \cline{2-8}
&D2 & 23.4 & 10.9 & 6.1 & 4.5 & 4.4 & 1\\ \cline{2-8}
&D3 & 32.2 & 16.1 & 9 & 0.3 & 0.5 & 0.2 \\ \cline{2-8}
&D4 & 5.4 & 2.7 & 1.7 & 12 & 10.7 & \hspace{0.6cm}$-$ \\ \cline{2-8}
&D5 & 2.8 & 1.8 & 1.4 & 0.8 & 1 & 1 \\ \hline
\end{tabular}
\end{center}
\caption{Performance comparison of AESVM (with $\epsilon = 10^{-3},10^{-4},10^{-5}$), CVM, BVM, LASVM and LIBSVM on datasets D1-D5}
\label{tb:testing1}
\end{table}

The results of the classification performance comparison on datasets D1-D5, are shown in Table \ref{tb:testing1}. It was observed that for all tested values of $\epsilon$, AESVM resulted in large reductions in training and classification times when compared to LIBSVM for a very small difference in classification accuracy. Most notably, for D3 the expected and overall training time speedups were of the order of $10^4$ and $10^3$ respectively, which is outstanding. Comparing the results of AESVM for different $\epsilon$ values, we see that $RMSE$ generally improves by decreasing when $\epsilon$ decreases, while the metrics improve by increasing when $\epsilon$ increases. The increase in $ETS$ and $OTS$ is of a larger order than the increase in $RMSE$ when $\epsilon$ increases. 

Comparing AESVM to CVM, BVM and LASVM, we see that AESVM in general gave the least values of $RMSE$ and the largest values of $ETS$, $OTS$, $ECS$ and $OCS$. In a few cases LASVM gave low $RMSE$ values. However, in all our experiments LASVM took longer to train than the other algorithms including LIBSVM. {\it We could not complete the evaluation of LASVM for D4 due to its large training time, which was more than 40 hours for some hyper-parameter combinations.} It was also found that LASVM sometimes resulted in a larger classification time than the other algorithms including LIBSVM. CVM and BVM generally gave high vales of $RMSE$. 

Table \ref{tb:testing1} compares the classification accuracy of CVM, BVM, LASVM and AESVM to the exact SVM solution given by LIBSVM. Another method to compare the algorithms is in terms of the maximum classification accuracy, and the mean and standard deviation of the classification accuracies, without using LIBSVM as a reference point. Such a comparison for datasets D1-D5, is given in Table \ref{tb:testing2}. The five algorithms under comparison were found to give similar maximum classification accuracies except for D2 and D4, where CVM and BVM gave significantly smaller values. Another interesting result is that for D3, the mean and standard deviation of accuracy of LASVM was found to be widely different from the other algorithms. For all the tested values of $\epsilon$ the maximum, mean and standard deviation of the classification accuracies of AESVM were found to be similar.

\begin{table}[h!]
\begin{center}
\begin{tabular}{|@{}p{1.57cm}!{\vrule width 1pt}@{}c@{}!{\vrule width 1pt}@{}p{1.5cm}|@{}p{1.5cm}|@{}p{1.5cm}|@{}p{1.4cm}|@{}p{1.4cm}|@{}p{1.57cm}|@{}p{1.4cm}|} \hline
Accuracy& Dataset& AESVM $\epsilon = 10^{-3}$ & AESVM $\epsilon = 10^{-4}$ & AESVM $\epsilon = 10^{-5}$ & CVM & BVM & LASVM &LIBSVM\\ \noalign{\hrule height 1pt}
\multirow{5}{1.9cm}{Maximum (x$10^2$)}&D1 & 93.4 & 93.8 & 93.6 & 94.1 & 94.4 & 94.3 &93.9\\ \cline{2-9}
&D2 & 77.1 & 77.2 & 77.8 & 70.3 & 67.1 & 78.1& 78.2\\ \cline{2-9}
&D3 & 99.9 & 99.9 & 99.9 & 99.9 & 99.9 & 99.9& 99.9 \\ \cline{2-9}
&D4 & 68.3 & 68.3 & 68.3 & 63.7 & 62.3 & \hspace{0.8cm}$-$ & 68.2\\ \cline{2-9}
&D5 & 98.7 & 98.8 & 98.9 & 99 & 99.1 & 99.2 & 99 \\ \noalign{\hrule height 1pt}
\multirow{5}{1.9cm}{Mean, standard deviation (x$10^2$)}&D1 & 92.2, 0.7 & 92.3, 0.8 & 92.3, 0.8 & 92.7, 0.8 & 92.6, 0.9 & 92.5, 0.8 & 92.4, 0.8\\ \cline{2-9}
&D2 & 72.3, 3.6 & 73.2, 3.7 & 73.6, 3.7 & 52.2, 0.8 & 54.6, 0.7 & 73.5, 0.5 & 74.1, 3.5\\ \cline{2-9}
&D3 & 99.8, 0 & 99.8, 0.1 & 99.8, 0.1 & 99.8, 0.2 & 99.8, 0.2 & 69.3, 29.9&99.8, 0.1\\ \cline{2-9}
&D4 & 61.3, 3.1 & 61, 3.1 & 61, 3.1 & 55.5, 3.1 & 54.9, 3.4 & \hspace{0.8cm}$-$ &60.6, 3.2\\ \cline{2-9}
&D5 & 96, 2.5 & 96.3, 2.6 & 96.5, 2.6 & 96.6, 2.5  & 97, 2 & 97, 2 & 96.6, 2.4 \\ \noalign{\hrule height 1pt}
\end{tabular}
\end{center}
\caption{Comparison of classification accuracies of AESVM (with $\epsilon = 10^{-3},10^{-4},10^{-5}$), CVM, BVM, LASVM and LIBSVM on datasets D1-D5}
\label{tb:testing2}
\end{table}

Next we present the results of performance comparison of CVM, BVM, LASVM, AESVM, and LIBSVM on the high-dimensional datasets D6-D9. As described in Section \ref{sec:derRS}, DeriveRS was run with only $\epsilon = 10^{-3}$ for these datasets. The results of the performance comparison are shown in Tables \ref{tb:testing3} and \ref{tb:testing4}. \emph{CVM was found to take longer than 40 hours to train on D6, D7 and D8 with some hyper-parameter values and hence we could not complete its evaluation for those datasets. BVM also took longer than 40 hours to train on D7 and it was also not evaluated for D7}. AESVM consistently reported $ETS$, $OTS$, $ECS$ and $OCS$ values that are larger than 1 unlike the other algorithms. Similar to the results in Table \ref{tb:testing1}, LASVM and BVM resulted in very large $RMSE$ values for some datasets. The results in Table \ref{tb:testing4} are similar to Table \ref{tb:testing2}, with similar maximum accuracies for all algorithms and significantly lower mean and higher standard deviation of accuracy for BVM and LASVM on some datasets.

\begin{table}[h!]
\begin{center}
\begin{tabular}{|c!{\vrule width 1pt}c!{\vrule width 1pt}p{1.5cm}|l|l|l|} \hline
Metric& Dataset& AESVM $\epsilon = 10^{-3}$ & CVM & BVM & LASVM\\ \noalign{\hrule height 1pt}
\multirow{4}{1.3cm}{$RMSE$ (x$10^2$)}&D6 & 0.21 & - & 7.8 & 0.85\\ \cline{2-6}
&D7 & 1.37 & - & - & 2.37 \\ \cline{2-6}
&D8 & 0.02 & - & 17.55 & 0  \\ \cline{2-6}
&D9 & 0.15 & 1 & 0.89 & 27.5 \\ \noalign{\hrule height 1pt}
\multirow{4}{*}{$ETS$}&D6 & 1.8 & - & 0.6 & 0.8\\ \cline{2-6}
&D7 & 1.4  & - & - & 0.9\\ \cline{2-6}
&D8 & 1.1 & - & 4.7 & 1 \\ \cline{2-6}
&D9 & 1.6 & 1.4 & 17.5 & 0.6 \\ \noalign{\hrule height 1pt}
\multirow{4}{*}{$OTS$}&D6 & 1.5 & - & 0.6 & 0.5\\ \cline{2-6}
&D7 & 1.2 & - & - & 0.7\\ \cline{2-6}
&D8 & 1.1 & - & 2.6 & 0.9 \\ \cline{2-6}
&D9 & 1.3 & 1.2 & 16.9 & 0.5 \\ \noalign{\hrule height 1pt}
\multirow{4}{*}{$ECS$}&D6 & 1.2 & - & 1.5 & 1\\ \cline{2-6}
&D7 & 1.16 & - & - & 1\\ \cline{2-6}
&D8 & 1 & - & 3.2 & 1 \\ \cline{2-6}
&D9 & 1.2 & 1.8 & 4.9 & 2.3 \\ \noalign{\hrule height 1pt}
\multirow{4}{*}{$OCS$}&D6 & 1.1 & - & 1.5 & 1\\ \cline{2-6}
&D7 & 1.1 & - & - & 1\\ \cline{2-6}
&D8 & 1 & - & 2.6 & 1 \\ \cline{2-6}
&D9 & 1.1 & 1.9 & 5.2 & 1.1 \\ \hline
\end{tabular}
\end{center}
\caption{Performance comparison of AESVM (with $\epsilon = 10^{-3}$), CVM, BVM, LASVM and LIBSVM on datasets D6-D9}
\label{tb:testing3}
\end{table}

\begin{table}[h!]
\begin{center}
\begin{tabular}{|@{}p{1.57cm}!{\vrule width 1pt}@{}c@{}!{\vrule width 1pt}@{}p{1.5cm}|@{}p{1.4cm}|@{}p{1.6cm}|@{}p{1.57cm}|@{}p{1.4cm}|} \hline
Accuracy& Dataset& AESVM $\epsilon = 10^{-3}$ & CVM & BVM & LASVM &LIBSVM\\ \noalign{\hrule height 1pt}
\multirow{4}{1.9cm}{Maximum (x$10^2$)}&D6 & 85.2 & - & 85.2 & 85 &85.1\\ \cline{2-7}
&D7 & 88.3 & - & - & 88.4 & 88.6 \\ \cline{2-7}
&D8 & 99.7 & - & 99.7 & 99.7 & 99.7 \\ \cline{2-7}
&D9 & 99.3 & 99.5 & 99.5 & 99.5 & 99.5 \\ \noalign{\hrule height 1pt}
\multirow{4}{1.9cm}{Mean, standard deviation (x$10^2$)}&D6 & 81.3, 2.8 & - & 80.2, 8.9 & 81.1, 2.9 & 81.4, 2.8\\ \cline{2-7}
&D7 & 85.3, 5.7 & - & - & 85.2, 6.2 & 85.7, 4.8\\ \cline{2-7}
&D8 & 92.3, 3.6 & - & 88.5, 18.1 & 92.3, 3.6 & 92.3, 3.6 \\ \cline{2-7}
&D9 & 98.7, 0.8 & 98.9, 0.8 & 98.9, 0.8 & 85.5, 23.9 & 98.8, 0.8 \\ \noalign{\hrule height 1pt}
\end{tabular}
\end{center}
\caption{Comparison of classification accuracies of AESVM (with $\epsilon = 10^{-3}$), CVM, BVM, LASVM and LIBSVM on datasets D6-D9}
\label{tb:testing4}
\end{table}

\subsubsection{Comparison to $\text{SVM}^{\mathrm{perf}}$}
$\text{SVM}^{\text{perf}}$ differs from the other SVM solvers in its ability to compute a solution close to the SVM solution for a given number of support vectors ($k$). The algorithm complexity is directly proportional to the parameter $k$, but with a decrease in $k$ the approximation becomes worse and the difference between the solutions of  $\text{SVM}^{\text{perf}}$ and SVM increases. We used a value of $k = 1000$ for our experiments, as it has been reported to give good performance \citep{Joachims09}. $\text{SVM}^{\text{perf}}$ was tested on datasets D1, D4, D5, D6, D8 and D9, with the Gaussian kernel \footnote{We used the software parameters '-t 2 -w 9 --i 2 --b 0 --k 1000' as suggested in the author's website} and the same hyper-parameter grid as described earlier. The results of the grid search are presented in Table \ref{tb:testing5}. The results of our experiments on AESVM (with $\epsilon = 10^{-3}$) and LIBSVM are repeated in Table \ref{tb:testing5} for ease of reference. The maximum, mean and standard deviation of classification accuracies are represented as max. Acc., mean Acc., and std. Acc. respectively.

\begin{table}[h!]
\begin{center}
\begin{tabular}{|@{}p{1.2cm}!{\vrule width 1pt}@{}p{1.6cm}@{}!{\vrule width 1pt}@{}p{1.2cm}|@{}l|@{}l|@{}l|@{}l|@{}p{1.8cm}|@{}p{1.8cm}|@{}p{1.6cm}|} \hline
Dataset & Solver & $RMSE$ (x$10^2$) & $ETS$ & $OTS$ & $ECS$ & $OCS$ & max. Acc. (x$10^2$)& mean Acc. (x$10^2$)& std. Acc. (x$10^2$)\\ \noalign{\hrule height 1pt}
\multirow{3}{1.9cm}{D1} &  AESVM & 0.28 & 451.5 & 92.1 & 4.8 & 3.8 & 93.4 & 92.2 & 0.7 \\ \cline{2-10}
&$\text{SVM}^{\text{perf}}$ & 0.74 & 3.7 & 0.9 & 6.8 & 6.8 & 94 & 92.7 & 0.5 \\ \cline{2-10}
& LIBSVM &   &  &  &  &  & 93.9 & 92.4 & 0.8\\ \noalign{\hrule height 1pt}
\multirow{3}{1.9cm}{D4} &  AESVM &  1.08 & 103.1 & 11.9 & 8.4 & 5.4 & 68.3 & 61.3 & 3.1 \\ \cline{2-10}
&$\text{SVM}^{\text{perf}}$ & 2.14 & 3.1 & 1.2 & 186.8 & 186.8 & 68.1 & 61.8 & 2.7 \\ \cline{2-10}
& LIBSVM &   &  &  &  & & 68.2 & 60.6 & 3.2 \\ \noalign{\hrule height 1pt}
\multirow{3}{1.9cm}{D5} &  AESVM &  0.99 & 40.2 & 5.2 & 4.3 & 2.8 & 98.7 & 96 & 2.5 \\ \cline{2-10}
&$\text{SVM}^{\text{perf}}$ &  0.26 & 0.2 & 0.1 & 5.8 & 5.8 & 99 & 96.7 & 2.4 \\ \cline{2-10}
& LIBSVM &  &  &  &  &  & 99 & 96.6 & 2.4\\ \noalign{\hrule height 1pt}
\multirow{3}{1.9cm}{D6} &  AESVM &  0.21 & 1.8 & 1.5 & 1.2 & 1.1 & 85.2 & 81.3 & 2.8\\ \cline{2-10}
&$\text{SVM}^{\text{perf}}$ &  9.39 & 1.1 & 0.9 & 20 & 20 & 85.2 & 79.6 & 10.7\\ \cline{2-10}
& LIBSVM &   &  &  & &  & 85.1 & 81.4 & 2.8\\ \noalign{\hrule height 1pt}
\multirow{3}{1.9cm}{D8} &  AESVM & 0.02 & 1.1 & 1.1 & 1 & 1 & 99.7 & 92.3 & 3.6\\ \cline{2-10}
&$\text{SVM}^{\text{perf}}$ & 54.2 & 37.6 & 23.8 & 49 & 49 & 99.9 & 55.7 & 42.3 \\ \cline{2-10}
& LIBSVM &  &  &  &  &  & 99.7 & 92.3 & 3.6\\ \noalign{\hrule height 1pt}
\multirow{3}{1.9cm}{D9} &  AESVM &  0.15 & 1.6 & 1.3 & 1.2 & 1.1 & 99.3 & 98.7 & 0.8\\ \cline{2-10}
&$\text{SVM}^{\text{perf}}$ & 22.6 & 1.2 & 0.9 & 21.3 & 21.3 & 99.2 & 86.1 & 18.8 \\ \cline{2-10}
& LIBSVM &   &  &  &  &  & 99.5 & 98.8 & 0.8\\ \noalign{\hrule height 1pt}
\end{tabular}
\end{center}
\caption{Performance comparison of $\text{SVM}^{\text{perf}}$, AESVM (with $\epsilon = 10^{-3}$), and LIBSVM}
\label{tb:testing5}
\end{table}

$\text{SVM}^{\text{perf}}$ was found to generally give higher $RMSE$ values than AESVM. In particular, for the high dimensional datasets (D6, D8 and D9), the $RMSE$ values were significantly higher. The training speedup values of $\text{SVM}^{\text{perf}}$ are much lower than AESVM except for D8. As expected, the classification time speedups of $\text{SVM}^{\text{perf}}$ are significantly higher than AESVM. The maximum accuracies of all the algorithms were similar. However, the mean and standard deviation of accuracies of $\text{SVM}^{\text{perf}}$ were very different from AESVM and LIBSVM for the high dimensional datasets D6, D8 and D9.

\subsubsection{Comparison to R$\mathrm{feat}$SVM} \label{sec:expRfeatSVM}
\citet{rahimi07} proposed a promising method to approximate non-linear kernel SVM solutions using simpler linear kernel SVMs. This is accomplished by first projecting the training dataset into a randomized feature space and then using any SVM solver with the linear kernel on the projected dataset.  We concentrated our experiments on investigating the accuracy of the solution of RfeatSVM and its similarity to the SVM solution. LIBSVM with the linear kernel was used to compute the RfeatSVM solution on the projected datasets. We used LIBSVM, in spite of the availability of faster linear SVM implementations, as it is an exact SVM solver. Hence only the performance metrics related to accuracy were used to compare the performance of AESVM, LIBSVM and RfeatSVM. The random Fourier features method, described in Algorithm 1 of \citet{rahimi07}, was used to project the datasets D1, D5, D6 and D9 into a randomized feature space of dimension E. The results of the accuracy comparison are given in Table \ref{tb:testing6}. We used a smaller hyper-parameter grid of all twenty four combinations of $C' = \{2^{-4},2^{-2},1,2^{2},2^{4},2^6\}$ and $g = \{2^{-4},2^{-2},1,2^2\}$ for our experiments. The results reported in Table \ref{tb:testing6} for AESVM and LIBSVM were computed for this smaller grid.

\begin{table}[h!]
\begin{center}
\begin{tabular}{|@{}p{1.2cm}|@{}p{2cm}@{}!{\vrule width 1pt}@{}p{1.3cm}|@{}p{1.8cm}|@{}p{1.8cm}|@{}p{1.6cm}|@{}p{1.6cm}|@{}p{2cm}|} \hline
Dataset & Solver & $RMSE$ (x$10^2$) & max. Acc. (x$10^2$)& mean Acc. (x$10^2$)& std. Acc. (x$10^2$) & Original density \% & Density after projection \%\\ \noalign{\hrule height 1pt}
\multirow{3}{1.9cm}{D1} &  AESVM & 0.24 & 93.6 & 92.2 & 0.9 & & \\ \cline{2-8}
& RfeatSVM (E = 100)& 56.18 & 37.8 & 36.1 & 1.3 & 33 & 100\\ \cline{2-8}
& LIBSVM &   & 93.6 & 92.3 & 0.9 & & \\ \noalign{\hrule height 1pt}
\multirow{3}{1.9cm}{D5} &  AESVM &  0.9 & 98.6 & 95.7 & 2.8 & & \\ \cline{2-8}
& RfeatSVM (E = 100) & 5.3 & 94.7 & 91.6 & 1.4 & 59& 100\\ \cline{2-8}
& LIBSVM &   & 98.9 & 96.2 & 2.7 & & \\ \noalign{\hrule height 1pt}
\multirow{3}{1.9cm}{D6} &  AESVM &  0.16 & 85.1 & 81.2 & 2.9 & & \\ \cline{2-8}
& RfeatSVM (E = 1000) &  4 & 81.6 & 78 & 2.2 & 11 & 100\\ \cline{2-8}
& LIBSVM &   & 85 & 81.3 & 3 & & \\ \noalign{\hrule height 1pt}
\multirow{3}{1.9cm}{D9} &  AESVM & 0.15 & 99.3 & 98.6 & 0.8 & & \\ \cline{2-8}
& RfeatSVM (E = 1000) & 0.6 & 98.7 & 97.4 & 0.6 & 4 & 95.8\\ \cline{2-8}
& LIBSVM &  & 99.5 & 98.8 & 0.9 & & \\ \noalign{\hrule height 1pt}
\end{tabular}
\end{center}
\caption{Performance comparison of RfeatSVM, AESVM (with $\epsilon = 10^{-3}$), and LIBSVM. The density of the datasets before and after projecting into randomized feature spaces are also shown}
\label{tb:testing6}
\end{table}

We used the same number of dimensions (E) of the randomized feature space for D1 and D6 as in \citet{rahimi07}. The $RMSE$ values for RfeatSVM were significantly higher than AESVM for most datasets, especially for D1 and D6. The maximum accuracy for RfeatSVM was found to be much less than AESVM and LIBSVM for all datasets. The time taken to compute the randomized feature space is not reported because it was found to be negligibly small. Another important observation was that the projected datasets were found to be almost 100\% dense. The training time of SVM solvers are typically linearly proportional to the density of the dataset and hence a highly dense dataset can take a significant training time even with fast linear SVMs. Dense datasets also have large memory requirements.

\subsection{Performance with the polynomial kernel} \label{sec:polyExp}

To validate our proposal of AESVM as a fast alternative to SVM for all non-linear kernels, we performed a few experiments with the polynomial kernel, $k({\bf x}_1,{\bf x}_2) = (1 + {\bf x}_1 ^T {\bf x}_2)^d$. The hyper-parameter grid composed of all twelve combinations of $C' = \{2^{-4},2^{-2},1,2^{2}\}$ and $d = \{2,3,4\}$ was used to compute the solutions of AESVM and LIBSVM on the datasets D1, D4 and D6. The results of the computation of the representative set using DeriveRS are shown in Table \ref{tb:testing7}. The parameters for DeriveRS were $P = 10^5$, $V= 10^3$ and  $\epsilon= 10^{-3}$, and the first level segregation was performed using FLS2. The performance comparison of AESVM and LIBSVM with the polynomial kernel is shown in Table \ref{tb:testing8}. Like in the case of the Gaussian kernel, we found that AESVM gave results similar to LIBSVM with the polynomial kernel, while taking shorter training and classification times.

\begin{table}[h!]
\begin{center}
\begin{tabular}{|@{}p{1.2cm}!{\vrule width 1pt}@{}c|@{}c|@{}c|}\hline
\multicolumn{4}{|c|}{$\frac{M}{N}$x100\%  (Computation time in seconds)}\\ \noalign{\hrule height 1pt}
Dataset&d = 2 &d = 3&d = 4\\ \hline\hline
D1 & 6.6(410) & 14.2(1329) & 22.5(3696)\\ \hline
D4 & 30.3(752) & 57.7(1839) & 76.5(2246)\\ \hline
D6 & 69(20) & 69.7(21) & 70.4(22)\\ \hline
\end{tabular}
\end{center}
\caption{Results of DeriveRS for the polynomial kernel}
\label{tb:testing7}
\end{table}

\begin{table}[h!]
\begin{center}
\begin{tabular}{|@{}p{1.2cm}!{\vrule width 1pt}@{}p{1.6cm}@{}!{\vrule width 1pt}@{}p{1.2cm}|@{}l|@{}l|@{}l|@{}l|@{}p{1.87cm}|@{}p{1.8cm}|@{}p{1.6cm}|} \hline
Dataset & Solver & $RMSE$ (x$10^2$) & $ETS$ & $OTS$ & $ECS$ & $OCS$ & max. Acc. (x$10^2$)& mean Acc. (x$10^2$)& std. Acc. (x$10^2$)\\ \noalign{\hrule height 1pt}
\multirow{2}{1.9cm}{D1} &  AESVM & 0.15 & 31.2 & 2 & 3.1 & 3.1 & 94 & 93.5 & 0.4 \\ \cline{2-10}
& LIBSVM &  &  &  &  &  & 94.1 & 93.5 & 0.4 \\ \noalign{\hrule height 1pt}
\multirow{2}{1.9cm}{D4} &  AESVM & 2.04 & 3.3 & 1.5 & 2 & 1.9 & 64.3 & 60.8 & 2.5 \\ \cline{2-10}
& LIBSVM &  &  &  &  &  & 64.5 & 60.7 & 2.5 \\ \noalign{\hrule height 1pt}
\multirow{2}{1.9cm}{D6} &  AESVM & 0.6 & 2.7 & 1.9 & 1.5 & 1.5 & 84.5 & 80.5 & 2.5 \\ \cline{2-10}
& LIBSVM &  &  &  &  &  & 84.6 & 81 & 2.3 \\ \noalign{\hrule height 1pt}
\end{tabular}
\end{center}
\caption{Performance comparison of AESVM (with $\epsilon = 10^{-3}$), and LIBSVM with the polynomial kernel}
\label{tb:testing8}
\end{table}

\section{Discussion} \label{sec:Disc}

\begin{figure}[h!]
\centering
\includegraphics[scale = 0.44]{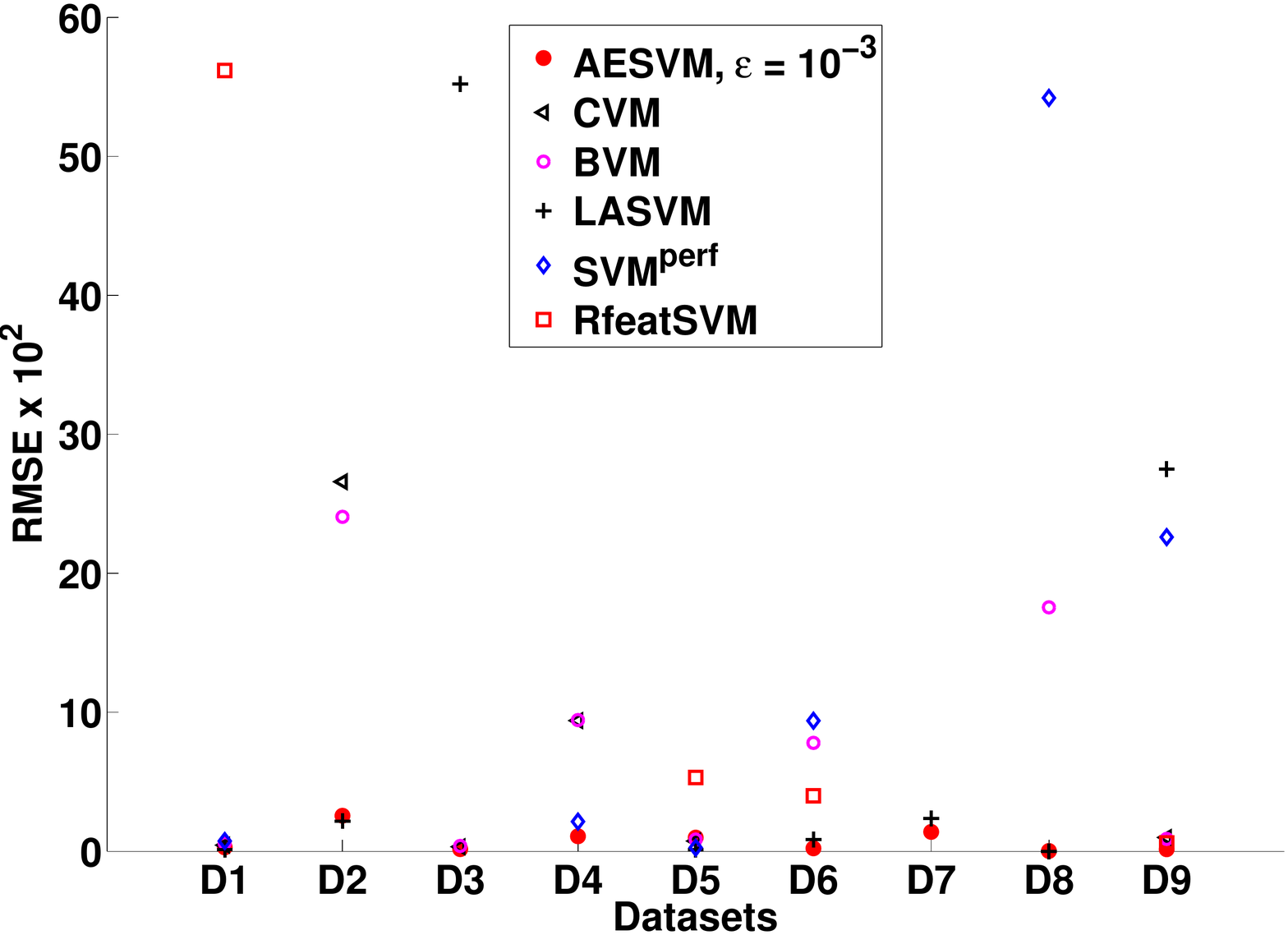}
\caption{Plot of RMSE values for all SVM solvers}\label{fig:resSummary1}
\end{figure}

\begin{figure}[h!]
\centering
\includegraphics[scale = 0.44]{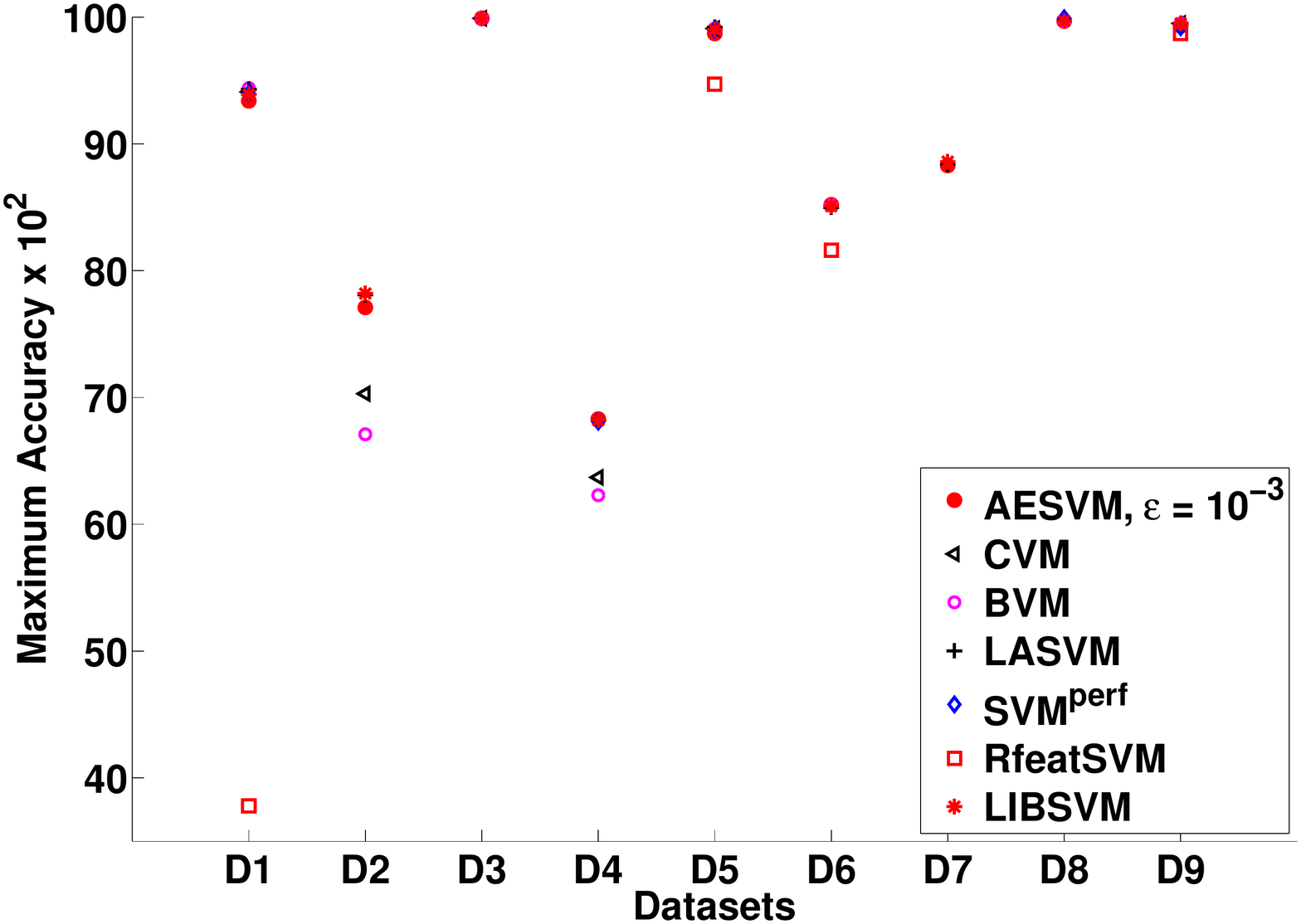}
\caption{Plot of maximum classification accuracy for all SVM solvers}\label{fig:resSummary2}
\end{figure}

\begin{figure}[h!]
\centering
\includegraphics[scale = 0.44]{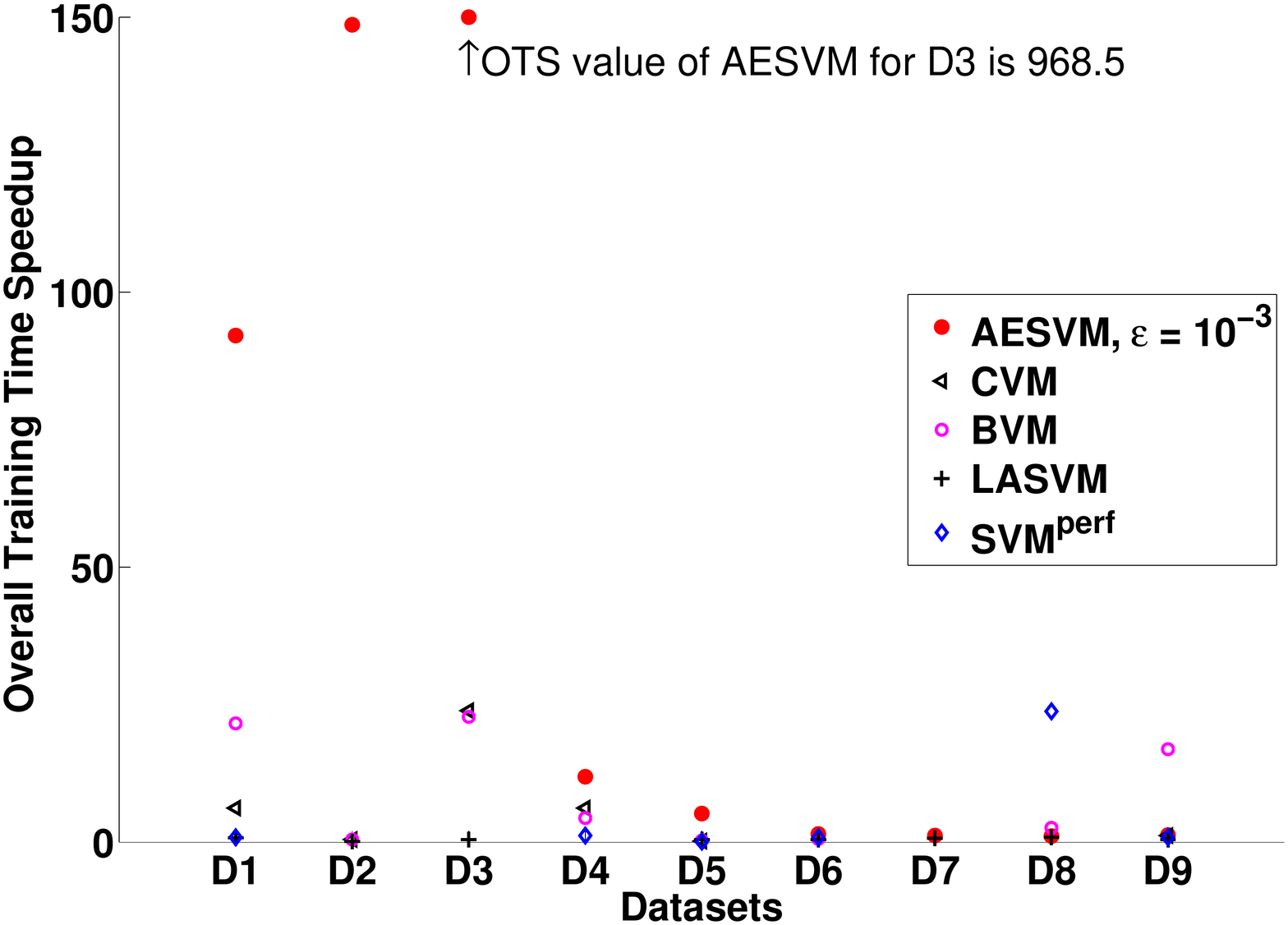}
\caption{Plot of overall training time speedup (compared to LIBSVM) for all SVM solvers}\label{fig:resSummary3}
\end{figure}

\begin{figure}[h!]
\centering
\includegraphics[scale = 0.44]{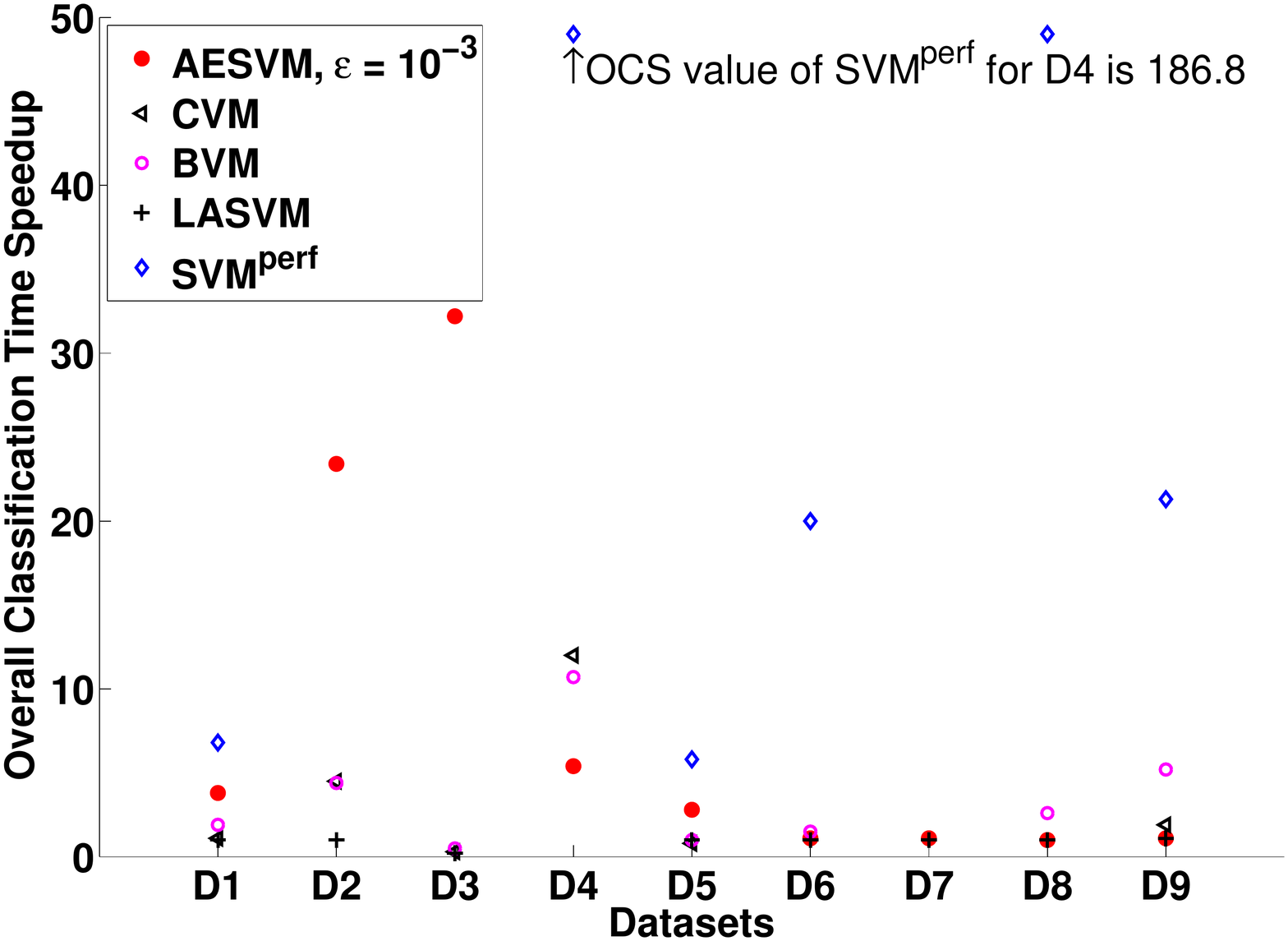}
\caption{Plot of overall classification time speedup (compared to LIBSVM) for all SVM solvers}\label{fig:resSummary4}
\end{figure}

AESVM is a new problem formulation that is almost identical to, but less complex than, the SVM primal problem. AESVM optimizes over only a subset of the training dataset called the representative set, and consequently, is expected to give fast convergence with most SVM solvers. In contrast, the other studies mentioned in Section \ref{sec:relWork} are mostly algorithms that solve the SVM primal or related problems. Methods such as RSVM also use different problem formulations. However, they require special algorithms to solve, unlike AESVM. In fact, AESVM can be solved using many of the methods in Section \ref{sec:relWork}. As described in Corollary 5, there are some similarities between AESVM and the Gram matrix approximation methods discussed earlier. It would be interesting to see a comparison of AESVM, with the core set based method proposed by \citet{Gartner09}. However, due to the lack of availability of a software implementation and of published results on L1-SVM with non-linear kernels using their approach, the authors find such a comparison study beyond the scope of this paper.

\emph{The theoretical and experimental results presented in this paper demonstrate that the solutions of AESVM and SVM are similar in terms of the resulting classification accuracy.} A summary of the experiments in Section \ref{sec:exp}, that compared an SMO based AESVM implementation, CVM, BVM, LASVM, LIBSVM, $\text{SVM}^{\text{perf}}$ and RfeatSVM, is presented in Figures \ref{fig:resSummary1} to \ref{fig:resSummary4}. \emph{It can be seen that AESVM typically gave the lowest approximation error ($RMSE$), while giving highest overall training time speedup ($OTS$). AESVM also gave competitively high overall classification time speedup ($OCS$) in comparison with the other algorithms except $\text{SVM}^{\text{perf}}$}. It was found that the maximum classification accuracies of all the algorithms except RfeatSVM were similar. RfeatSVM, and in some cases CVM and BVM, gave lower maximum classification accuracies. Though the results of RfeatSVM illustrated in Figures \ref{fig:resSummary1} and \ref{fig:resSummary2}, were computed for a smaller hyper-parameter grid (refer Section \ref{sec:expRfeatSVM}), we believe it indicates the overall performance of the method. Apart from the excellent experimental results for AESVM with the Gaussian kernel, AESVM also gave good results with the polynomial kernel as described in Section \ref{sec:polyExp}. 

The algorithm DeriveRS was generally found to be efficient, especially for the lower dimensional datasets D1-D5. For the high dimensional datasets D6-D9, the representative set was almost the same size as the training dataset, resulting in small gains in training and classification time speedups for AESVM. In particular, for D8 (MNIST dataset) the representative set computed by DeriveRS was almost 100\% of the training set. A similar result was reported for this dataset in \citet{Beygelzimer06}, where a divide and conquer method was used to speed up nearest neighbor search. Dataset D8 is reported to have resulted in nearly no speedup, compared to a speedup of almost one thousand for other datasets when their method was used. Their analysis found that the data vectors in D8 were very distant from each other in comparison with the other datasets \footnote{This is indicated by the large expansion constant for D8 illustrated in \citet{Beygelzimer06}}. This observation can explain the performance of DeriveRS on D8, as data vectors that are very distant from each other are expected to have large representative sets. It should be noted that irrespective of the dimensionality of the datasets, AESVM always resulted in excellent performance in terms of classification accuracy. There seems to be no relation between dataset density and the performance of DeriveRS and AESVM.

The authors will provide the software implementation of AESVM and DeriveRS upon request. Based on the presented results, we suggest the parameters $\epsilon = 10^{-3}$, $P = 10^5$ and $V = 10^3$ for DeriveRS. A possible extension of this paper is to apply the idea of the representative set to other SVM variants and to support vector regression (SVR). It is straightforward to see that the theorems in Section \ref{sec:prop} can be extended to SVR. It would be interesting to investigate AESVM solvers implemented using methods other than SMO. Modifications to DeriveRS using the methods in Section \ref{sec:relWork} might improve its performance on high dimensional datasets. The authors will investigate improvements to DeriveRS and the application of AESVM to the linear kernel in their future work.

\acks{Dr. Khargonekar acknowledges support from the Eckis professor endowment at the University of Florida. Dr. Talathi was partially supported by the Children's Miracle Network, and the Wilder Center of Excellence in Epilepsy Research. The authors acknowledge Mr. Shivakeshavan R. Giridharan, for providing assistance with computational resources.}

\end{document}